\documentclass[nonacm,sigconf]{acmart}
\usepackage{hyperref}

\AtBeginDocument{%
  }

\usepackage{comment}
\usepackage{hhline}
\usepackage{multirow}
\usepackage{xspace}

\usepackage[acronym, nonumberlist]{glossaries}
\newacronym{ai}{AI}{Artificial Intelligence}
\newacronym{ml}{ML}{Machine Learning}
\newacronym{nlp}{NLP}{Natural Language Processing}
\newacronym{nli}{NLI}{Natural Language Inference}
\newacronym{mnli}{MNLI}{Multi-Genre Natural Language Inference}
\newacronym{vlm}{VLM}{Vision-Language Model}
\newacronym{vqa}{VQA}{Visual Question Answering}
\newacronym{se}{SE}{Semantic Entropy}
\newacronym{vase}{VASE}{Vision-Amplified Semantic Entropy}
\newacronym{hedge}{HEDGE}{Hallucination Estimation via Dense Geometric Entropy}
\newacronym{uq}{UQ}{Uncertainty Quantification}
\newacronym{auc}{AUC}{Area Under the Curve}
\newacronym{roc}{ROC}{Receiver Operating Characteristic}
\newacronym{rocauc}{ROC-AUC}{Receiver Operating Characteristic Area Under the Curve}
\newacronym{clip}{CLIP}{Contrastive Language-Image Pre-training}
\newacronym{llm}{LLM}{Large Language Model}

\newacronym{cnn}{CNN}{Convolutional Neural Network}
\newacronym{vit}{ViT}{Vision Transformer}
\newacronym{lstm}{LSTM}{Long Short-Term Memory}
\newacronym{rnn}{RNN}{Recurrent Neural Network}
\newacronym{nn}{NN}{Neural Network}
\newacronym{cv}{CV}{Computer Vision}

\glsdisablehyper

\usepackage{listings}
\usepackage{array}
\makeglossaries

\usepackage{xcolor,soul}
\sethlcolor{lightgray}

\begin{document}

\title[VideoHEDGE]{VideoHEDGE: Entropy-Based Hallucination Detection for Video-VLMs via Semantic Clustering and Spatiotemporal Perturbations}

\author{Sushant Gautam}
\affiliation{%
  \institution{SimulaMet \& OsloMet}
  \city{Oslo}
  \country{Norway}
}
\email{sushant@simula.no}

\author{Cise Midoglu}
\affiliation{%
  \institution{Forzasys}
  \city{Oslo}
  \country{Norway}
}
\email{cisemidoglu@gmail.com}

\author{Vajira Thambawita}
\affiliation{%
  \institution{SimulaMet}
  \city{Oslo}
  \country{Norway}
}
\email{vajira@simula.no}

\author{Michael A. Riegler}
\affiliation{%
  \institution{Simula Research Laboratory}
  \city{Oslo}
  \country{Norway}
}
\email{michael@simula.no}

\author{Pål Halvorsen}
\affiliation{%
  \institution{SimulaMet, Forzasys \& OsloMet}
  \city{Oslo}
  \country{Norway}
}
\email{paalh@simula.no}

\begin{abstract}
Hallucinations in video-capable vision-language models (Video-VLMs) remain frequent and high-confidence, while existing uncertainty metrics often fail to align with correctness. We introduce VideoHEDGE, a modular framework for hallucination detection in video question answering that extends entropy-based reliability estimation from images to temporally structured inputs. Given a video-question pair, VideoHEDGE draws a baseline answer and multiple high-temperature generations from both clean clips and photometrically and spatiotemporally perturbed variants, then clusters the resulting textual outputs into semantic hypotheses using either Natural Language Inference (NLI)-based or embedding-based methods. Cluster-level probability masses yield three reliability scores: Semantic Entropy (SE), RadFlag, and Vision-Amplified Semantic Entropy (VASE). We evaluate VideoHEDGE on the SoccerChat benchmark using an LLM-as-a-judge to obtain binary hallucination labels. Across three 7B Video-VLMs (Qwen2-VL, Qwen2.5-VL, and a SoccerChat-finetuned model), VASE consistently achieves the highest ROC-AUC, especially at larger distortion budgets, while SE and RadFlag often operate near chance. We further show that embedding-based clustering matches NLI-based clustering in detection performance at substantially lower computational cost, and that domain fine-tuning reduces hallucination frequency but yields only modest improvements in calibration. The \texttt{hedge-bench} PyPI library enables reproducible and extensible benchmarking, with full code and experimental resources available at \url{https://github.com/Simula/HEDGE\#videohedge} .
\end{abstract}

\maketitle
\section{Introduction}

Vision--language models (VLMs) are rapidly becoming the default interface for reasoning about rich visual data such as images, videos, and interactive environments. Recent video-capable VLMs promise to answer natural-language questions about complex, temporally extended events, from sports broadcasts to surveillance footage, using a single, unified interface. However, as these models move closer to deployment in safety- and decision-critical settings, a persistent failure mode remains: they hallucinate. Models frequently produce fluent, high-confidence answers that are inconsistent with the underlying video, while offering little internal signal that anything has gone wrong.

Existing work on uncertainty estimation for VLMs has focused largely on image-based visual question answering (VQA) and other static settings. Typical approaches rely on token-level entropy, confidence over class labels, or heuristic consistency checks across multiple samples. While these methods can perform adequately for image-based VQA, they struggle in two key ways when extended to Video-VLMs. First, they ignore temporal structure and the fact that video inputs admit a continuum of spatiotemporal distortions (for example, fewer frames, lower resolution, or additive noise) that may selectively degrade visual evidence. Second, they treat answers as isolated strings rather than elements of a semantic hypothesis space in which many surface forms can describe the same underlying event. As a result, uncertainty metrics often fail to distinguish grounded predictions from hallucinations, especially when models are overconfident.

Recent work on HEDGE has shown that combining answer sampling, semantic clustering, and entropy over clusters can yield better reliability estimates for image-based VQA. Directly lifting this framework to video, however, leaves several open challenges. Video inputs introduce a much larger sampling space over frames, resolutions, and distortions, and naive extensions of logical clustering via Natural Language Inference (NLI) become prohibitively expensive at realistic sampling scales. Moreover, it is unclear how best to exploit spatiotemporal perturbations, such as photometric shifts and noise, that preserve the underlying event while stressing the model's visual grounding.

In this work, we introduce VideoHEDGE, a modular framework for hallucination detection in video question answering that generalizes entropy-based reliability estimation from images to temporally structured inputs. Given a video--question pair, VideoHEDGE draws a baseline answer and multiple high-temperature generations from both clean clips and photometrically and spatiotemporally perturbed variants. These answers are then clustered into semantic hypotheses, using either NLI-based or embedding-based methods, and cluster-level probability masses are used to define reliability metrics. Crucially, VideoHEDGE treats perturbations as controlled probes of visual robustness: by contrasting semantic distributions induced by clean versus degraded videos, it can detect answers whose validity depends on fragile or spurious visual cues.

We instantiate VideoHEDGE on the SoccerChat benchmark~\cite{gautam2025soccerchat} for soccer video understanding, using Qwen3-30B-A3B~\cite{Qwen3Yang2025May} as an LLM adjudicator to obtain binary hallucination labels. Across three 7B Video-VLMs, a general-purpose Qwen2-VL-instruct model~\cite{Qwen2-VL}, an improved Qwen2.5-VL-instruct model~\cite{Qwen25vl}, and SoccerChat-qwen2-vl-7b\cite{gautam2025soccerchat}, we compare three entropy-based scores: Semantic Entropy (SE), RadFlag, and Vision-Amplified Semantic Entropy (VASE). We further contrast NLI-based clustering with a more scalable embedding-based alternative, and study how reliability behaves under varying frame counts, pixel budgets, and perturbation strengths. Our experiments show that domain fine-tuning substantially reduces hallucination frequency but yields only modest gains in calibration, and that perturbation-aware metrics like VASE offer the most consistent improvements in hallucination detection.

Our contributions are threefold:
\begin{itemize}
    \item \textbf{Framework:} We propose VideoHEDGE, a modular reliability-estimation pipeline for Video-VLMs that unifies spatiotemporal sampling, semantic clustering, and entropy-based scoring, built on top of the open-source \texttt{hedge-bench} package.
    \item \textbf{Metrics and analysis:} We adapt and evaluate three cluster-level metrics, SE, RadFlag, and VASE, for video VQA, and show that VASE, which explicitly exploits clean--noisy semantic gaps, consistently achieves the highest ROC--AUC for hallucination detection, particularly at larger distortion budgets.
    \item \textbf{Scalability and empirical insights:} We demonstrate that embedding-based clustering matches NLI-based clustering in detection performance at far lower computational cost, and we characterize how architecture, domain fine-tuning, and spatiotemporal resolution jointly shape both hallucination rates and the effectiveness of reliability metrics.
\end{itemize}

\section{Related Work}
\label{sec:related}

\textbf{Hallucination in (multimodal) foundation models.} 
Hallucination, fluent model outputs that contradict underlying evidence, is a well-documented failure mode of large language models (LLMs). Recent surveys provide unified taxonomies distinguishing intrinsic vs.\ extrinsic hallucinations and factual vs.\ reasoning errors, and analyze contributing factors in data, training objectives, and decoding dynamics~\cite{ACMSurvey2025}. 
Several works extend these definitions to multimodal settings, emphasizing that visual or audio grounding failures may coexist with linguistically coherent responses~\cite{survey_lvml_hallucination_liu_2024, survey_lvml_hallucination_sahoo_2024, MMSurvey2025, AHESurvey2024}. 
Our formulation adopts this broader notion and focuses specifically on hallucinations where generated answers disagree with the \emph{video} evidence despite high model confidence.

\textbf{Hallucination in large vision--language models.}
Dedicated analyses of hallucinations in large vision--language models (LVLMs) catalog object, attribute, relational, and cross-modal hallucinations~\cite{survey_lvml_hallucination_lan_2024, ObjectHalBench2024, THRONE2024}. 
These works highlight that most evaluations are image-centric, rarely considering temporal dependencies or perturbation sensitivity. In contrast, hallucinations in Video-VLMs may arise from misinterpreting motion, event outcomes, or temporal ordering. Moreover, existing benchmarks rarely exploit controlled spatiotemporal perturbations. This gap motivates our work: VideoHEDGE develops hallucination detection \emph{for Video-VLMs} under systematically varied frame and pixel budgets.

\textbf{Video question answering and video--language understanding.}
Video question answering (VideoQA) is a foundational task for video--language reasoning, with recent surveys reviewing architectures, datasets, and temporal reasoning challenges~\cite{survey_videoqa_2024, survey_video_language_understanding_2025}. Broader reviews of video--language models (Vid-LLMs) summarize captioning, dialog, temporal localization, and long-video reasoning tasks~\cite{survey_vidllm_2024, survey_multimodal_llms_2025}. 
Despite extensive work on model architecture and pretraining, evaluation overwhelmingly focuses on task accuracy (e.g., answer correctness, caption similarity), and reliability estimation remains largely unaddressed. 
Our approach is orthogonal: VideoHEDGE assumes a fixed Video-VLM and estimates reliability by analyzing semantic behavior under controlled sampling and perturbations.

\textbf{Sports video understanding.}
Sports video understanding, most prominently for association football (soccer), has become a structured domain for evaluating event-centric video reasoning. The SoccerNet benchmark family~\cite{soccernet_v1, soccernet_v2, Giancola2022Oct} provides broadcast soccer videos with annotations for action spotting, replay grounding, and temporal reasoning. 
More recently, X-VARS~\cite{xvars_2024} introduces a referee-oriented multimodal LLM capable of event captioning and rule-based reasoning on soccer clips. 
While these systems improve prediction quality and domain adaptation, they do not provide explicit hallucination detection or reliability signals. 
VideoHEDGE complements such models by adding a \emph{model-agnostic reliability layer} specialized for short-clip queries in the SoccerChat-style~\cite{gautam2025soccerchat}.

\textbf{Uncertainty quantification and calibration for multimodal LLMs.}
Uncertainty quantification (UQ) for LLMs has been surveyed extensively, covering probabilistic calibration, abstention, and self-consistency methods~\cite{uq_llms_survey_2025, calibration_llms_survey_2024}. 
For multimodal LLMs, training-free UQ frameworks such as UMPIRE~\cite{umpire_2025} combine sample diversity with internal coherence measures to estimate error likelihood across text, image, and video tasks. 
Other work explores grounding-based calibration and consistency-based reliability for LVLMs~\cite{llm_vlm_calibration_2024}. 
These approaches typically operate on token-level entropy or scalar confidence signals. 
In contrast, VideoHEDGE models uncertainty at the \emph{semantic cluster} level and explicitly analyzes how semantic distributions shift under spatiotemporal perturbations, connecting to semantic-dispersion methods such as Semantic Entropy~\cite{SE2024}, Semantic Entropy Probes~\cite{Kossen2024SEP}, multimodal semantic robustness~\cite{VASE2025, DSE2025}, and perturbation-based UQ~\cite{gao2024spuq, padhi2025multimodal, epistemic-uq2024}.

\textbf{Semantic entropy and entropy-based hallucination detection.}
Our work builds on semantic-entropy-based detection methods. Farquhar et al.\ propose \textit{semantic entropy}, which clusters multiple generations into semantic equivalence classes and computes entropy over these clusters, strongly correlating with correctness in text tasks~\cite{SE2024}. 
Subsequent work explores cheaper approximations and domain-specific adaptations~\cite{Kossen2024SEP}. 
In vision--language settings, Vision-Amplified Semantic Entropy (VASE) has been applied to medical VQA, comparing semantic beliefs across clean and perturbed images~\cite{VASE2025}. 
HEDGE~\cite{HEDGE} systematizes these ideas for \emph{image} VQA, incorporating answer sampling, NLI- and embedding-based clustering, and metrics such as Semantic Entropy, RadFlag~\cite{RadFlag2024}, and VASE. 
VideoHEDGE extends this decomposition to \textit{videos}, addressing temporally structured sampling, spatiotemporal perturbations, and the large candidate sets induced by frame/pixel variations.

\textbf{Benchmarks and mitigation methods for LVLM hallucinations.}
Several works benchmark or mitigate hallucinations in LVLMs. THRONE~\cite{THRONE2024} provides an object-level hallucination benchmark for free-form image descriptions, while multi-object hallucination datasets~\cite{mo_hallucination_2024} analyze entity-level inconsistencies. 
Mitigation methods such as Visual Contrastive Decoding (VCD)~\cite{vcd_2024} and retrieval-augmented decoding~\cite{contrastive_retrieval_2024} reduce hallucinations during generation. 
These approaches modify the model or its decoding procedure. 
VideoHEDGE takes a complementary direction: it does \emph{not} alter the underlying Video-VLM, but provides post-hoc, perturbation-aware reliability estimates driven by semantic clustering.

\textbf{Semantic clustering for reliability estimation.}
Semantic entropy and HEDGE employ NLI-based entailment graphs to form logically coherent clusters, but at quadratic computational cost~\cite{HEDGE}. 
Recent work shows that sentence-embedding-based clustering offers a scalable approximation for high-throughput settings~\cite{Abdaljalil2025Mar}. 
In parallel, LLM-as-a-judge protocols have become standard for labeling hallucination quality~\cite{THRONE2024}. 
VideoHEDGE integrates these components: binary hallucination labels are provided by an external LLM judge, while reliability metrics use NLI- or embedding-based clustering. 
Empirically, we show that embedding-based clustering achieves comparable ROC--AUC to NLI-based clustering while enabling large spatiotemporal perturbation budgets in the video setting.

\section{Methodology}
\label{sec:methodology}

\begin{figure*}[htb]
    \includegraphics[width=0.95\textwidth]{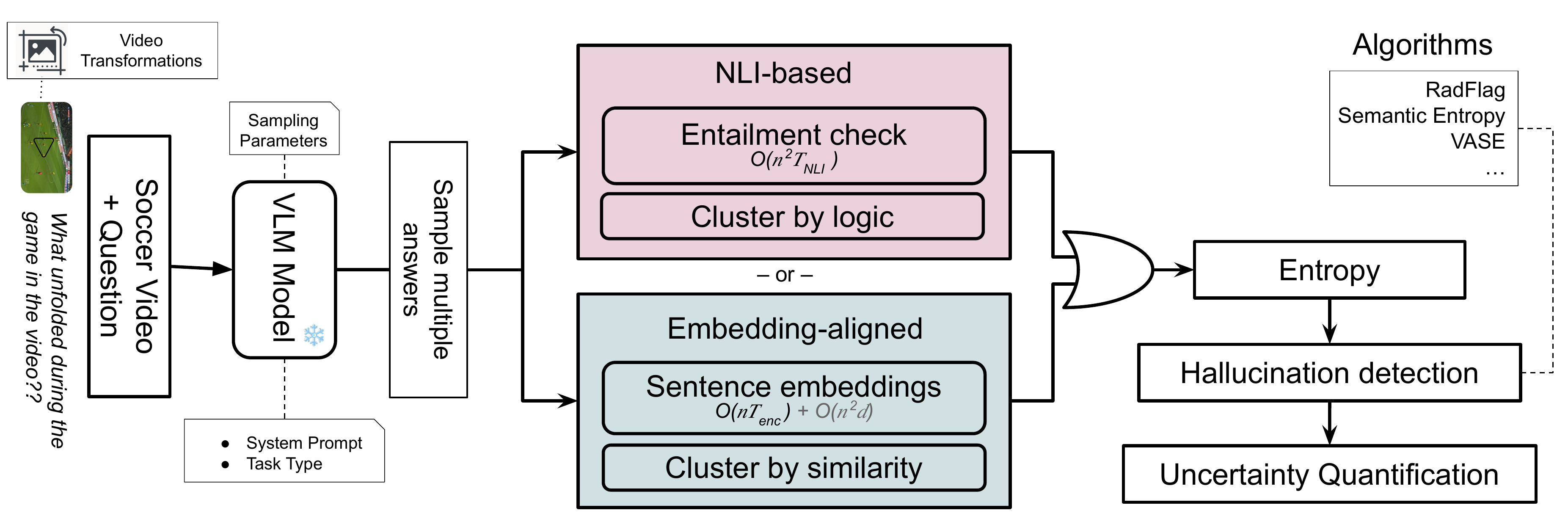}
    \centering
    \caption{Overview of the proposed VideoHEDGE framework for hallucination detection in visual question answering (VQA).
    A vision--language model generates multiple answers per video--question pair, which are grouped via two strategies:
    natural language inference (NLI)-based logical clustering and embedding-aligned semantic clustering.
    Entropy within these groups quantifies uncertainty, enabling hallucination detection through metrics such as
    RadFlag, Semantic Entropy, and VASE.}
    \label{fig:teaser}
\end{figure*}

\Gls{hedge} was originally proposed as a modular framework for estimating reliability in image-based VQA, combining answer sampling, semantic clustering, and entropy-based scoring into a single pipeline~\cite{HEDGE}. In this work we extend this idea to videos and introduce \emph{VideoHEDGE}, a variant tailored to temporally structured inputs and soccer-domain evaluation. Rather than designing new metrics from scratch, VideoHEDGE instantiates the same high-level decomposition---\emph{sampling} $\rightarrow$ \emph{clustering} $\rightarrow$ \emph{scoring}---but (i) adapts the sampling stage to video clips, (ii) treats temporal and spatial perturbations as controlled sources of visual degradation, and (iii) implements scalable clustering and metric computation for the large candidate sets common in video VQA.

Concretely, VideoHEDGE standardizes how we obtain multiple answers from a video--question pair under different frame and resolution settings, how we group these answers into semantic clusters using either \gls{nli}-based or embedding-based backends, and how we compute reliability scores with off-the-shelf metrics (RadFlag, \gls{se}, \gls{vase}) as well as their embedding-aligned variants. By viewing reliability estimation as a robustness problem over the geometry of semantic clusters, VideoHEDGE enables consistent comparison across video-capable multimodal models. We implement our framework on top of the open-source \texttt{hedge-bench} package, which we extend with video-specific sampling and perturbation modules.

\subsection{Pipeline Overview}
\label{subsec:pipeline_overview}

Figure~\ref{fig:teaser} summarizes the VideoHEDGE pipeline. For each video--question pair, the framework (i) generates multiple answers under controlled sampling conditions, (ii) arranges them into a sequence with associated log-likelihoods, (iii) performs semantic clustering in either an NLI or embedding space, and (iv) aggregates probability mass at the cluster level to derive reliability scores. The video is only used in the sampling stage; all subsequent steps operate purely on the textual answers.

Given a video $v$ and question $q$, we first obtain a low-temperature baseline answer $A_0$ that approximates the model's most likely response. We then draw high-temperature samples under two visual conditions. \emph{Clean} answers $A_1,\dots,A_n$ are generated from the reference video using a fixed frame-selection policy (e.g., a 24-frame uniform subsample at a chosen resolution). \emph{Noisy} answers $N_1,\dots,N_n$ are generated from photometrically perturbed variants of the same clip.

For each noisy variant, we sample brightness $b \sim \mathcal{U}(-0.2, 0.2)$, contrast $c \sim \mathcal{U}(0.8, 1.2)$, saturation $s \sim \mathcal{U}(0.95, 1.05)$, and a hue shift
$h \sim \mathcal{U}(-0.02, 0.02)\cdot 360^\circ$, and apply these jointly together with additive spatiotemporal noise to all frames via a video-processing pipeline~\cite{ffmpeg}. This produces stochastic color- and noise-perturbed videos $V'$ that preserve the underlying event while degrading visual evidence. In our experiments (Section~\ref{sec:experiments}), we vary the number of such noisy samples systematically and match the clean sampling budget to maintain balance.
We record the mean token log-likelihood for every generated answer. We denote the resulting arrays by $\log \ell_{\text{clean}}$ and $\log \ell_{\text{noisy}}$ for clean and noisy generations, respectively. The exact number of trials $n$ depends on the sampling and distortion budgets of the experimental setting.

\subsection{Semantic Clustering}
\label{subsec:nli_vs_embedding}

VideoHEDGE converts the sequence of textual answers into a discrete set of semantic hypotheses by clustering them. The video is not used in this stage. Each response is mapped to a cluster identifier $c_i = \text{cluster}(A_i)$ (or $\text{cluster}(N_i)$), with indices following the sequence layout:
$
[0]\!\rightarrow\!A_0,\quad
[1{:}n{+}1]\!\rightarrow\!A_1..A_n,\quad
[n{+}1{:}2n{+}1]\!\rightarrow\!N_1..N_n.
$
We compare two strategies that trade off semantic precision and computational cost.

\textbf{NLI-based clustering.} 
Following~\cite{HEDGE}, clustering is treated as a logical consistency problem. Given a set $S=\{t_1,\dots,t_m\}$, a \gls{mnli}-style model predicts a label $y_{ij}\in\{\text{entails}\allowbreak\,\text{contradicts},\text{neutral}\}$ for each pair $(t_i,t_j)$. We build a directed entailment graph $G_{\text{ent}}$ with edges: 
$
E_{\text{ent}} = \{(i,j)\mid y_{ij} = \text{entails}\},
$
and define a mutual-entailment relation $i \sim j$ if and only if $(i,j)\in E_{\text{ent}}$ and $(j,i)\in E_{\text{ent}}$. Clusters are obtained as the transitive closure of this relation via a union--find procedure, with contradiction links used to prevent merging of answers that explicitly disagree.
This strategy offers fine-grained logical control but requires $\mathcal{O}(n^{2} T_{\text{NLI}})$ model evaluations, where $T_{\text{NLI}}$ is the cost of a single NLI forward pass, making it expensive for large sampling scales or many perturbation settings.

\textbf{Embedding-based clustering.} 
To improve scalability, we also cluster in a continuous embedding space. Each text $t_i$ is encoded into a sentence vector $\mathbf{x}_i \in \mathbb{R}^d$ using a pretrained encoder~\cite{reimers2019sentence}, then normalized to unit length. Pairwise cosine similarities are given by $\sigma_{ij} = \mathbf{x}_i^\top \mathbf{x}_j$.
We build an undirected similarity graph $G_{\text{sim}}$ where vertices are responses and edges connect pairs that are either above a similarity threshold or within the local neighborhood:
$
\sigma_{ij} \ge \tau \quad \text{or} \quad j \in \text{kNN}(i).
$
Clusters correspond to connected components in $G_{\text{sim}}$. The threshold $\tau$ is tuned on a validation split to maximize \gls{rocauc} for hallucination detection~\cite{fawcett2006ROC}, giving automatic control over cluster granularity.

The complexity of this approach is
$\mathcal{O}(n T_{\text{enc}}) + \mathcal{O}(n^{2} d)$, where $T_{\text{enc}}$ is the cost of a single encoder pass. While quadratic in $n$, the $\mathcal{O}(n^{2} d)$ term consists solely of dot products and graph operations, which are substantially cheaper than full NLI model evaluations. This makes embedding-based clustering well-suited to the larger candidate pools that arise when we vary frame counts, pixel budgets, or distortion types in VideoHEDGE.

\subsection{Hallucination Metrics for Video VQA}
\label{subsec:metrics}

Given the semantic clusters and log-likelihoods from the sampling stage, VideoHEDGE computes reliability scores by aggregating probability mass at the cluster level and then measuring uncertainty and robustness.

\subsubsection{Semantic distribution and entropy}

Let $\log \ell_i$ be the mean token log-likelihood of answer $i$ and $c_i$ its cluster assignment. We define a normalized semantic distribution over clusters as
$
s_j =
\frac{
    \exp\!\left(\sum_{i: c_i = j} \exp(\log \ell_i - \max_k \log \ell_k)\right)
}{
    \sum_{m} \exp\!\left(\sum_{i: c_i = m} \exp(\log \ell_i - \max_k \log \ell_k)\right)
},
\label{eq:semantic-distribution}
$
and the corresponding semantic entropy,
$
\mathrm{SE}(\mathbf{s})
= -\sum_{j} s_j \log s_j.
\label{eq:semantic-entropy}
$
Following Shannon's classical formulation of entropy~\cite{Shannon1948}, this quantity measures uncertainty over semantic clusters rather than over token-level outputs.

We compute separate distributions for clean and noisy generations, $\mathbf{s}_{\text{clean}}$ and $\mathbf{s}_{\text{noisy}}$, respectively. %
The difference between these distributions reflects how sensitive the model’s semantic beliefs are to temporal or spatial perturbations in the input video clip.

\subsubsection{Semantic Entropy (SE)}

\Gls{se} quantifies uncertainty over clusters induced by clean generations:
$
\mathrm{SE} = \mathrm{SE}(\mathbf{s}_{\text{clean}}).
\label{eq:se}
$
High values indicate that probability mass is spread across many competing semantic explanations for the same video, a signal that the model may be under-specified or relying on unstable cues.

\subsubsection{RadFlag}

RadFlag measures how often high-temperature clean answers leave the baseline cluster:
$
\mathrm{RadFlag}
= 1 - \frac{1}{n}\sum_{i=1}^{n}\mathbf{1}[c_i = c_0],
\label{eq:radflag}
$
where $c_0$ is the cluster of the baseline answer $A_0$ and $c_i$ are those of the clean samples $A_i$. Values near~1 indicate that the model frequently jumps to semantically different answers when sampling from the same video, even without explicit input noise.

\subsubsection{Vision-Amplified Semantic Entropy (VASE)}

\Gls{vase} extends \gls{se} by explicitly incorporating the gap between clean and noisy semantic distributions~\cite{VASE2025}. We first form a shifted distribution that accentuates clusters whose probability changes under visual degradation:
$
\mathrm{VASE} =
\mathrm{SE}\!\left(
\operatorname{softmax}\!\left(
\mathbf{s}_{\text{clean}}
+
\alpha(\mathbf{s}_{\text{clean}} - \mathbf{s}_{\text{noisy}})
\right)
\right),
\label{eq:vase}
$
with a scaling factor $\alpha$ controlling sensitivity to the clean--noisy semantic gap (we set $\alpha = 1$ in all experiments following \cite{VASE2025}). Intuitively, VASE is high when the model’s semantic beliefs are both uncertain and unstable under changes to the video, such as fewer frames or lower resolution.

\subsubsection{Effect of distortion budget}

Increasing the number of visual distortions enlarges the pool of noisy answers, and we proportionally increase the clean sampling budget to maintain balance. As a consequence, all three metrics, $\mathrm{SE}$, $\mathrm{RadFlag}$, and $\mathrm{VASE}$, become functions of both sampling scale and perturbation strength. When we vary the sampling scale in our experiments, the resulting trends should therefore be interpreted as the outcome of this coupled design.

All metrics are implemented in a batched manner using stable mean-log-probability aggregation, enabling efficient evaluation over large video datasets and multiple perturbation settings.

\subsection{Task-Specific Prompt Templates}
\label{subsec:answer_length}

VideoHEDGE does not vary linguistic verbosity on the SoccerChat dataset~\cite{gautam2025soccerchat}; instead, it relies on task-specific prompt templates aligned with the two annotation regimes of the dataset. The underlying video is fixed, and only the task description and expected answer type differ between \emph{EventClassification} and \emph{VideoQA}.

For \emph{EventClassification}, the system prompt describes the model as a soccer video assistant that must identify the single most salient football event in the clip (e.g., goal, penalty, corner kick, foul, or no event). The user message provides the short clip and a generic query asking what key event occurs. The expected answer is a concise event description, typically a single categorical label or a short phrase that can be mapped to the dataset’s event taxonomy.

For \emph{VideoQA}, we reuse the same high-level system instructions but present the clip together with a task-specific natural language question. These questions are drawn from the SoccerChat annotations and target aspects such as the acting player, the outcome of a shot, or a brief description of what happens. The model is allowed to respond in free-form natural language, but the prompts emphasize concise, directly grounded answers.

Both templates are held fixed across all models and perturbation settings. This design ensures that differences in the reliability metrics of Section~\ref{subsec:metrics} can be attributed to model behavior, spatiotemporal conditions, and clustering choices, rather than to changes in answer-length constraints or stylistic instructions.

\begin{figure}[htb]
\centering
\lstset{
    basicstyle=\ttfamily\scriptsize,
    breaklines=true,
    breakatwhitespace=false,
    columns=fullflexible,
    keepspaces=true,
    showstringspaces=false,
    frame=none,
}
\begin{lstlisting}[caption={Prompt templates for each answer-length prompt configuration. The system prompt is omitted for SoccerChat-qwen2-vl-7b where its default is used.}, label={lst:task_prompts}]
{"EventClassification": [{"role": "system", "content": 
    "You are a sports video reasoning assistant. Given a short video clip and a user question, provide an answer that is concise and directly addresses exactly what is asked. Ground the answer strictly in the video content. When referring to teams or players, always use jersey colors. Do not give explanations or extra text."}, {"role": "user", "content": 
    "<video> Identify the key event shown in the clip."}],
"VideoQA": [{"role": "system", "content":
    "You are a sports video reasoning assistant. Given a short video clip and a user question, provide an answer that is concise and directly addresses exactly what is asked. Ground the answer strictly in the video content. When referring to teams or players, always use jersey colors. Do not give explanations or extra text."}, {"role": "user", "content": 
"<video> {r.question}" }]}
\end{lstlisting}
\end{figure}

\begin{figure}[htb]
\centering
\lstset{
    basicstyle=\ttfamily\scriptsize,
    breaklines=true,
    breakatwhitespace=false,
    columns=fullflexible,
    keepspaces=true,
    showstringspaces=false,
    frame=none,
}
\begin{lstlisting}[caption={Evaluation prompt template for both EventClassification and VideoQA tasks used with Qwen3-30B-A3B adjudicator.}, label={lst:evaluation_prompts}]
[{"role": "system", "content": 
"You are a fair and careful evaluator of sports video question-answer pairs, focused on football (soccer). You will be given: 
    - question: the exact text shown to the model 
    - description: optional clarification 
    - correct_answer: the verified correct answer 
    - generated_answer: the model's answer. 
    Task type: {task_type}. 
- Video is ALWAYS football. 
If task_type == "EventClassification":
- Question: identify the single most relevant event. correct_answer and generated_answer are SHORT LABELS (e.g., goal, penalty, foul, etc.). Labels count as MATCHING if they indicate the SAME event. Score 0 when event types differ or one is "no event" and the other is an actual event. 
If task_type == "VideoQA": 
- Answers must be short and directly address the question. Team names vs jersey colors are equivalent if they refer to the same side. Paraphrasing allowed unless contradictory. 
Scoring: Score 1 if generated_answer conveys the same main fact/event. Score 0 if different, contradicted, or missing. Output (STRICT JSON, no code fences): {"reason": "<one concise sentence (less than 20 words)>", "score": 0 or 1}"}, {
"role": "user", "content": 
    "task_type: {task_type}; 
    question: {question}; 
    description: {description}; 
    correct_answer: {correct_answer};
    generated_answer: {generated_answer}"}]
\end{lstlisting}
\end{figure}

Listings \ref{lst:task_prompts} and \ref{lst:evaluation_prompts} describe the task prompts used for model generation and the adjudication prompt used by \texttt{Qwen3-30B-A3B}. Under the default SoccerChat inference setting (24 frames, 100,352 max pixels), the adjudicator reveals substantial variation in grounding across models. The base \texttt{Qwen2-VL} produces predominantly hallucinated responses, with 2,578 answers deemed unsupported and only 342 deemed evidence-aligned. \texttt{Qwen2.5-VL} shows a modest improvement (2,461 unsupported vs. 459 supported), while the fine-tuned \texttt{SoccerChat-qwen2-vl} model demonstrates a markedly stronger grounding profile: 1,885 hallucinated predictions and 1,035 supported ones, making it the only model whose supported outputs exceed its hallucinated outputs.

\section{Experiments}
\label{sec:experiments}

We instantiate VideoHEDGE for reliability evaluation on soccer video understanding tasks. All experiments use the pipeline introduced in Section~\ref{sec:methodology}, integrating spatiotemporal sampling, semantic clustering, and the metrics defined in Section~\ref{subsec:metrics}. We evaluate multiple Video-Language Models (VLMs) under controlled temporal and spatial perturbations and compare three reliability scores across two task types: \emph{EventClassification} and \emph{VideoQA}.

\subsection{Dataset and Task Setup}
\label{subsec:dataset}

We use the \textbf{SoccerChat} dataset, a curated collection of short soccer video clips paired with human-authored queries~\cite{gautam2025soccerchat}. Each clip contains between 3 and 6 seconds of gameplay and is associated with two task families. In \emph{EventClassification}, the model must identify the single most salient football event (e.g., \emph{goal}, \emph{penalty}, \emph{corner kick}, \emph{foul}, \emph{no event}). In \emph{VideoQA}, the model answers a task-specific question about the clip, such as identifying the acting player, describing the outcome of a shot, or summarizing what happens. For computational efficiency, we work on a randomly selected subset of the SoccerChat dataset consisting of 490 clips that have a single annotated event, resulting in 1,460 (video, question, answer) instances.

These two task types allow us to examine reliability under structured (classification-like) and open-ended (free-form QA) settings using the same underlying videos.

\subsection{Evaluated Models}
\label{subsec:models}

We evaluate three 7B-parameter VLMs chosen to disentangle architectural gains from domain specialization. \textbf{Qwen2-VL-7B-Instruct} is a general-purpose VLM supporting video inputs and multimodal reasoning. \textbf{Qwen2.5-VL-7B-Instruct} is a newer generation with an upgraded vision encoder and more efficient attention stack. \textbf{SoccerChat-qwen2-vl-7B} is a domain-specialized model obtained by fine-tuning Qwen2-VL-7B on the SoccerChat dataset for fine-grained soccer understanding. This selection reveals whether reliability improvements arise primarily from architectural changes (Qwen2~$\rightarrow$~Qwen2.5) or from domain exposure (generalist~$\rightarrow$~SoccerChat).

\subsection{Video Sampling and Perturbation Setup}
\label{subsec:sampling_setup}

For each video-question pair, we generate one low-temperature \emph{baseline} response $A_0$, $n$ \emph{clean} high-temperature samples $\{A_1,\dots,A_n\}$, and $n$ \emph{noisy} samples $\{N_1,\dots,N_n\}$ from visually perturbed videos $V'$, as described in Sec.~\ref{subsec:pipeline_overview}. Clean responses are generated from the original clip using a default 24-frame uniform sampling policy unless stated otherwise. Every clean or noisy response is paired with its mean token log-likelihood for subsequent cluster-level scoring.

Noisy variants $V'$ are created using the photometric perturbation function introduced earlier:
$
\text{brightness} \sim \mathcal{U}(-0.2, 0.2),\quad
\text{contrast} \sim \mathcal{U}(0.8, 1.2),\quad
\text{sat.} \sim \mathcal{U}(0.95, 1.05),\quad
\text{hue shift} \sim \mathcal{U}(-0.02, 0.02)\cdot 360^\circ,
$
combined with spatiotemporal noise via the FFmpeg \texttt{noise} filter~\cite{ffmpeg}. These perturbations modify the video’s color and noise distribution while leaving the underlying event intact.

\subsection{Temporal and Spatial Sensitivity Conditions}
\label{subsec:sensitivity}

To study how VLMs depend on spatiotemporal detail, we vary two input dimensions. First, we evaluate each model at frame counts $\{4, 8, 12, 16, 20, 24, 30\}$ per clip; the default SoccerChat fine-tuning uses 24 frames. Second, we vary the maximum per-frame pixel budget,
$
10{,}000,\; 40{,}000,\; 100{,}352,\; 160{,}000,\,
$
and 
$
250{,}000,
$
corresponding approximately to resolutions from $128\times72$ up to $656\times369$. These controlled conditions allow VideoHEDGE to probe whether failures arise due to insufficient temporal evidence, insufficient spatial detail, or model-specific brittleness.

In all cases, the clean and noisy sampling budgets grow with the distortion budget, so the metrics in Section~\ref{subsec:metrics} should be interpreted as functions of both perturbation strength and sampling scale.

\subsection{Adjudication of Ground-Truth Labels}
\label{subsec:adjudication}

Binary reliability labels (\texttt{0 = hallucinated}, \texttt{1 = supported}) are obtained via an LLM-as-a-judge setup. We employ Qwen3-30B-A3B with a task-specific evaluation prompt (Listing~2). The adjudicator receives the task type (EventClassification or VideoQA), the question, the reference (gold) answer, and the model-generated answer, and returns a JSON object
$
\{\text{score}: 0\text{ or }1,\; \text{reason}: \text{string}\}.
$

Scoring criteria differ across task types. For EventClassification, only agreement on the underlying event is accepted (e.g., ``goal'' vs.\ ``header leading to goal'' is considered a match, but ``goal'' vs.\ ``saved shot'' is scored as~0). For VideoQA, paraphrasing is permitted as long as the meaning matches the reference answer. This procedure yields consistent, high-quality labels for thousands of model outputs.

\subsection{Evaluation Protocol}
\label{subsec:evaluation}

From the semantic clusters and log-likelihoods, we compute three reliability scores: Semantic Entropy (SE)~\cite{SE2024}, RadFlag~\cite{RadFlag2024}, and VASE~\cite{VASE2025}, as defined in Section~\ref{subsec:metrics}. For each score, we compare two clustering backends. The default backend uses embedding-based clustering with normalized MiniLM SentenceTransformer embeddings (\href{https://huggingface.co/sentence-transformers/all-MiniLM-L6-v2}{all-MiniLM-L6-v2})~\cite{reimers2019sentence,wang2020minilm}. The clustering threshold~$\tau$ is selected to maximize SE performance on the VideoQA task, and the same $\tau$ is then used to compute all other metrics for that setting.
 For comparison, we also report results with NLI-based clustering using a DeBERTa MNLI model~\cite{DeBERTaHe2020Jun}, as in Section~\ref{subsec:nli_vs_embedding}.

Our primary evaluation metric is ROC-AUC between each reliability score and the adjudicator-provided binary labels. This quantifies how well each score separates hallucinated vs.\ supported responses for a given model, task type, and perturbation condition.

\subsection{Runtime and Scalability}
\label{subsec:runtime}
\begin{table*}[htb]
\centering
\caption{
ROC AUC of hallucination detection for SoccerChat-qwen2-vl-7b model across varying numbers of maximal distortions (1 to 10) for EventClassification and VideoQA tasks. Each block reports SE, RadFlag, and VASE metrics for both embedding- and NLI-based clustering. Bold highlighted entries denote the best-performing configuration for each metric within a block. Visualized in Figure~\ref{fig:auc_vs_distortion}.
}
\label{tab:distortion_variations}
\begin{tabular}{lcccccccccc}
\toprule
 & \multicolumn{10}{c}{\textbf{Clustering by Embedding}} \\
\cmidrule(lr){2-11}
\textbf{Max Distortions} & \textbf{1} & \textbf{2} & \textbf{3} & \textbf{4} & \textbf{5} & \textbf{6} & \textbf{7} & \textbf{8} & \textbf{9} & \textbf{10} \\
\midrule
\multicolumn{11}{l}{\textbf{EventClassification}} \\
SE       & 0.500 & 0.527 & 0.503 & 0.517 & 0.532 & 0.517 & 0.536 & 0.554 & 0.515 & 0.513 \\
RadFlag  & 0.521 & 0.538 & 0.500 & 0.515 & 0.523 & 0.513 & 0.521 & 0.546 & 0.514 & 0.516 \\
VASE     & 0.569 & 0.609 & 0.617 & 0.648 & 0.662 & 0.654 & 0.663 & \hl{\textbf{0.671}} & 0.656 & 0.654 \\
\midrule
\multicolumn{11}{l}{\textbf{VideoQA}} \\
SE       & 0.500 & 0.516 & 0.509 & 0.518 & 0.521 & 0.528 & 0.515 & 0.520 & 0.531 & 0.524 \\
RadFlag  & 0.504 & 0.506 & 0.507 & 0.509 & 0.513 & 0.526 & 0.522 & 0.503 & 0.523 & 0.523 \\
VASE     & 0.537 & 0.560 & 0.594 & 0.606 & 0.612 & 0.618 & 0.620 & 0.610 & 0.621 & \hl{\textbf{0.623}} \\
\midrule
\\[-0.8em]
 & \multicolumn{10}{c}{\textbf{Clustering by NLI}} \\
\cmidrule(lr){2-11}
\textbf{Max Distortions} & \textbf{1} & \textbf{2} & \textbf{3} & \textbf{4} & \textbf{5} & \textbf{6} & \textbf{7} & \textbf{8} & \textbf{9} & \textbf{10} \\
\midrule
\multicolumn{11}{l}{\textbf{EventClassification}} \\
SE       & 0.500 & 0.512 & 0.520 & 0.513 & 0.526 & 0.511 & 0.514 & 0.515 & 0.514 & 0.532 \\
RadFlag  & 0.492 & 0.505 & 0.508 & 0.505 & 0.509 & 0.506 & 0.508 & 0.510 & 0.510 & 0.531 \\
VASE     & 0.538 & 0.576 & 0.588 & 0.597 & 0.597 & 0.597 & 0.607 & 0.616 & 0.617 & \hl{\textbf{0.635}} \\
\midrule
\multicolumn{11}{l}{\textbf{VideoQA}} \\
SE       & 0.500 & 0.518 & 0.515 & 0.515 & 0.511 & 0.517 & 0.511 & 0.519 & 0.520 & 0.514 \\
RadFlag  & 0.509 & 0.511 & 0.509 & 0.505 & 0.507 & 0.511 & 0.510 & 0.512 & 0.513 & 0.512 \\
VASE     & 0.546 & 0.584 & 0.593 & 0.600 & 0.607 & 0.620 & 0.621 & 0.627 & 0.626 & \hl{\textbf{0.631}} \\
\bottomrule
\end{tabular}
\end{table*}

\begin{figure}[htb]
\includegraphics[width=0.5\textwidth]{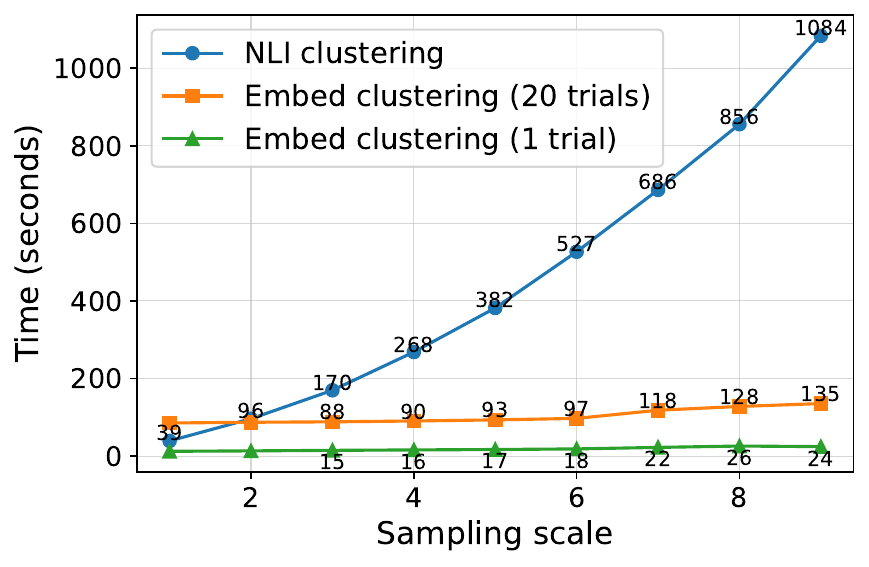}
    \centering
 \caption{
Computational scalability of NLI-based and embedding-based clustering methods for reliability estimation, evaluated using the SoccerChat model on the SoccerChat dataset. 
}
\label{fig:scale_performance}
\end{figure}

To assess scalability, we measure GPU time and memory while varying the distortion budget, i.e., the number of answer samples per clip (see Figure \ref{fig:scale_performance}). Embedding-based clustering exhibits near-flat scaling, with per-sample runtimes in the range of 12 to 135\,s even in our largest setting. In contrast, NLI-based clustering grows quadratically and exceeds 1{,}000\,s and 3.8\,GB of VRAM at 9 distortions. These measurements establish embedding-based clustering as the only practical choice for video-level perturbation analysis at scale, while still retaining NLI-based clustering as a high-precision reference.

\subsection{Reproducibility}
\label{subsec:reproducibility}

All experiments employ cached embeddings and log-likelihoods, and batched inference for both VLMs and the adjudicator. We release intermediate artifacts (answers, and labels), enabling reproducible computation of all scores across models, perturbation settings, and sampling scales.

\section{Results}
\label{sec:results}

We evaluate VideoHEDGE by measuring how well automated reliability scores align with hallucination labels produced by the Qwen3-30B-A3B adjudicator. Concretely, we report (i) hallucination vs.\ support counts under different spatiotemporal settings and (ii) ROC-AUC between reliability scores and binary labels across models, clustering backends, and perturbation budgets. All conclusions below are drawn directly from the quantitative results reported in Tables~\ref{tab:distortion_variations} and~\ref{tab:hedge_video_auc}.

\begin{table*}[htb]
\centering
\caption{
ROC-AUC comparison of hallucination detection methods on the HEDGE-Video SoccerChat benchmark.
Results are reported for EventClassification and VideoQA tasks across three model outputs:
Qwen2-VL-7B-Instruct, Qwen2.5-VL-7B-Instruct, and SoccerChat-Qwen2VL-FT.
Bold highlighted entries indicate the best-performing method within each configuration.
} 
\label{tab:hedge_video_auc}
\setlength{\tabcolsep}{6pt}
\renewcommand{\arraystretch}{1.2}

\begin{tabular}{lcccccccccccc}
\toprule
\multirow{3}{*}{\textbf{Model}} &
\multicolumn{6}{c}{\textbf{Event Classification}} &
\multicolumn{6}{c}{\textbf{VideoQA}} \\
\cmidrule(lr){2-7} \cmidrule(lr){8-13}
& \multicolumn{3}{c}{\textbf{Embedding}} 
& \multicolumn{3}{c}{\textbf{NLI}} 
& \multicolumn{3}{c}{\textbf{Embedding}} 
& \multicolumn{3}{c}{\textbf{NLI}} \\
\cmidrule(lr){2-4} \cmidrule(lr){5-7}
\cmidrule(lr){8-10} \cmidrule(lr){11-13}
& SE & RadFlag & VASE & SE & RadFlag & VASE
& SE & RadFlag & VASE & SE & RadFlag & VASE \\
\midrule
\textbf{Qwen2-VL}
& 0.464 & 0.485 & 0.470  %
& 0.495 & 0.504 & \hl{\textbf{0.577}}  %
& 0.550 & 0.548 & \hl{\textbf{0.624}}  %
& 0.558 & 0.555 & 0.630 \\        %
\textbf{Qwen2.5-VL}
& 0.627 & 0.611 & \hl{\textbf{0.644}}
& 0.618 & 0.593 & 0.621
& 0.579 & 0.593 & 0.587
& 0.595 & 0.579 & \hl{\textbf{0.640}} \\
\textbf{SoccerChat\tiny{-qwen2-vl}}
& 0.513 & 0.516 & \hl{\textbf{0.654}}
& 0.532 & 0.531 & 0.635
& 0.524 & 0.523 & 0.623
& 0.514 & 0.512 & \hl{\textbf{0.631}} \\
\bottomrule
\end{tabular}
\end{table*}

\textbf{Reliability profile across models.}
Under the default SoccerChat configuration (24 frames, 100{,}352 max pixels), the adjudicated labels indicate that all three evaluated VLMs still produce a substantial number of unsupported answers. Qwen2-VL is judged to yield 2{,}578 answers with score~0 (unsupported) versus 342 with score~1 (supported), while Qwen2.5-VL improves this modestly to 2{,}461 unsupported and 459 supported answers. In contrast, the domain-adapted SoccerChat-qwen2-vl model yields a more balanced outcome, with 1{,}381 unsupported and 1{,}539 supported answers across both tasks. This suggests that domain adaptation has a much stronger effect on correctness than the architectural change from Qwen2 to Qwen2.5. However, these adjudicated counts alone do not reveal whether unreliable answers are detectable from the models’ own uncertainty patterns; this is precisely what the proposed reliability metrics aim to assess.

\textbf{Unimodal uncertainty vs.\ vision-amplified scoring.}
Table~\ref{tab:hedge_video_auc} summarizes ROC-AUC scores for hallucination detection across models, tasks, and clustering backends. For the base Qwen2-VL model on EventClassification, Semantic Entropy (SE) and RadFlag perform close to random chance: SE achieves 0.464 (embedding) and 0.495 (NLI), while RadFlag is 0.485 and 0.504. Similar near-chance behaviour appears for several other settings, indicating that hallucinations in video VQA can be highly confident and internally consistent over clean samples, making clean-only uncertainty signals unreliable. This “failure” of unimodal metrics is not universal, however: for Qwen2.5-VL on EventClassification, SE reaches 0.627 (embedding) and 0.618 (NLI), with RadFlag at 0.611 and 0.593, suggesting that architectural improvements can make clean-only uncertainty somewhat more informative in certain regimes.

Across most models and tasks, Vision-Amplified Semantic Entropy (VASE) provides the strongest or near-strongest separation between hallucinated and supported answers. For Qwen2-VL, VASE reaches 0.577 on EventClassification (NLI) and 0.624--0.630 on VideoQA, clearly outperforming SE and RadFlag in the corresponding blocks. For Qwen2.5-VL, VASE attains 0.644 (EventClassification, embedding) and 0.640 (VideoQA, NLI). For SoccerChat-qwen2-vl, VASE achieves the highest reported scores, with 0.654 on EventClassification (embedding) and 0.631 on VideoQA (NLI). While the strongest VASE configurations lie in the moderate 0.57--0.67 range, their consistent advantage supports the core design choice of VideoHEDGE: incorporating the divergence between clean and visually perturbed generations provides a more informative reliability signal than clean-only uncertainty.

\textbf{Embedding vs.\ NLI clustering.}
The comparison between clustering backends reveals that embedding-based clustering can often match NLI-based clustering in practice, despite being much cheaper. In Table~\ref{tab:hedge_video_auc}, VASE under embedding and NLI backends is particularly close for the SoccerChat-qwen2-vl model (within 0.02 AUC), though gaps can be larger for the generic Qwen models. There are also cases where NLI is noticeably better (e.g., Qwen2-VL EventClassification, 0.577 vs.\ 0.470), but there is no systematic advantage in either direction. Given that sentence embeddings already capture much of the semantic geometry for short answers and event labels, these results suggest that embedding-based clustering is sufficient for hallucination detection on SoccerChat at the sampling scales considered here.

Runtime measurements make this trade-off explicit. As the distortion budget increases, NLI-based clustering exhibits steep growth in GPU time, exceeding 1{,}000\,s and 3.8\,GB VRAM at 9 distortions, whereas embedding-based clustering remains in the 12 to 135\,s range for the tested sample sizes. Together with the near-parity in ROC-AUC for SoccerChat, this makes embedding-based clustering the only practical option for large perturbation budgets, while NLI-based clustering is best viewed as a high-precision but resource-intensive reference.

\begin{figure}[htb]
    \centering
\includegraphics[width=0.5\textwidth]{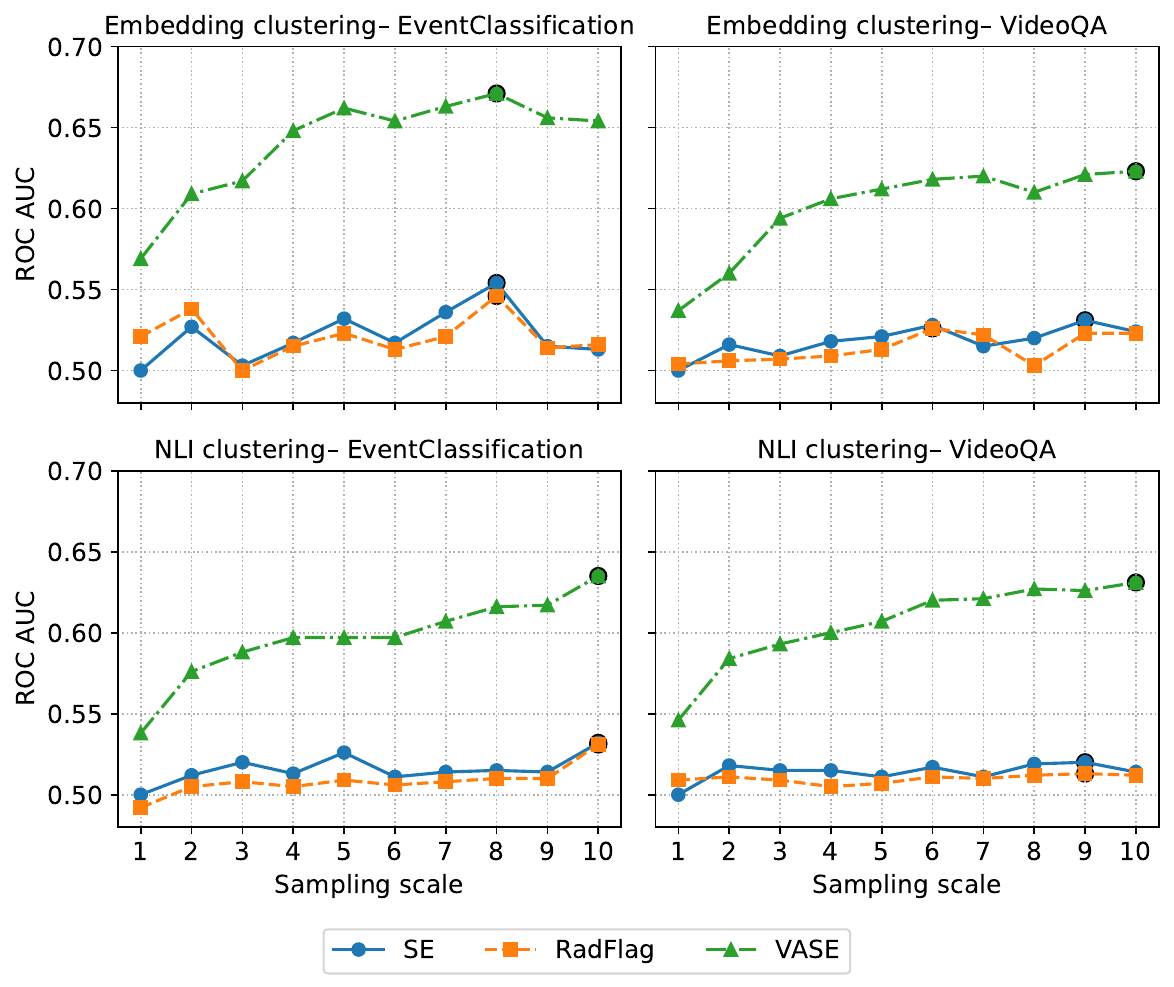}
    \centering
\caption{
Effect of sampling scale and the associated visual perturbations on hallucination-detection performance for SoccerChat-qwen2-vl evaluated on the SoccerChat dataset.
ROC AUC scores for three metrics (Semantic Entropy, RadFlag, and VASE) evaluated on two tasks: EventClassification and VideoQA. Values reported in Table~\ref{tab:distortion_variations}.
}
\label{fig:auc_vs_distortion}   
\end{figure}

\textbf{Effect of distortion budget.} 
Table~\ref{tab:distortion_variations} examines how reliability scores evolve as we increase the maximal number of distortions from 1 to 10 for SoccerChat-qwen2-vl. For both EventClassification and VideoQA tasks, and for both clustering backends, VASE generally improves when moving from a single distortion to a larger ensemble, even though the progression is not strictly monotonic. For example, under embedding-based clustering on VideoQA, VASE increases from 0.537 (1 distortion) to 0.594 (3 distortions) and eventually to 0.623 (10 distortions), with a small dip around 8 distortions. Under NLI clustering, VASE on VideoQA grows from 0.546 (1 distortion) to 0.631 (10 distortions). In contrast, SE and RadFlag fluctuate mildly around chance (roughly 0.50--0.56) across the entire sweep, showing little sensitivity to the perturbation budget.

These trends indicate that the main benefits of larger distortion budgets accrue specifically to perturbation-aware metrics like VASE, which rely on differences between clean and noisy semantic distributions. A small number of noisy samples provides only a weak signal, but once 6 to 10 perturbations are available, the semantic drift induced by visual degradation becomes strong enough to noticeably improve discrimination between hallucinated and supported answers. We stop short of identifying an exact optimal budget, since the curves are noisy, but the best AUCs in Table~\ref{tab:distortion_variations} consistently occur in the high-distortion regime.

\textbf{Sensitivity to temporal and spatial resolution.}
We next study how hallucination behaviour and detection performance vary with the number of frames and the maximum pixel budget per frame for SoccerChat-qwen2-vl using embedding-based clustering. Table~\ref{tab:framescores} shows that the fraction of supported answers is relatively stable as we vary frame counts between 4 and 30: for EventClassification, supported counts range from 860 to 893; for VideoQA, from 631 to 654. Similarly, Table~\ref{tab:maxpixels} indicates that very low resolutions (10{,}000 pixels) lead to many more hallucinations than supports, while higher resolutions (100{,}352 pixels and above) flip this ratio in favour of supported answers for EventClassification and substantially reduce it for VideoQA.

\begin{table}[htb]
\centering
\caption{Score distributions (0 = hallucinated, 1 = supported) for EventClassification and VideoQA across frame-count settings. Evaluations performed by \texttt{Qwen3-30B-A3B}.}
\label{tab:framescores}
\begin{tabular}{lccccc}
\multirow{2}{*}{\textbf{Frames}} & \multirow{2}{*}{\textbf{FPS}} & \multicolumn{2}{c}{\textbf{EventClassification}} & \multicolumn{2}{c}{\textbf{VideoQA}} \\
 &  & \textbf{0} & \textbf{1} & \textbf{0} & \textbf{1} \\
\hline
4  & 0.4 & 600 & 860 & 806 & 654 \\
8  & 0.8 & 580 & 880 & 808 & 652 \\
12 & 1.2 & 590 & 870 & 817 & 642 \\
16 & 1.6 & 567 & 893 & 825 & 635 \\
20 & 2.0 & 584 & 876 & 814 & 646 \\
\hl{\textbf{24}} & 2.4 & 567 & 893  &814 & 646  \\
30 & 3.0 & 585 & 875 & 829 & 631 \\
\hline
\end{tabular}
\end{table}

\begin{table}[htb]
\centering
\small
\caption{Score distributions (0 = hallucinated, 1 = supported) for EventClassification and VideoQA across different max-pixels settings. Evaluations performed by \texttt{Qwen3-30B-A3B}.}
\label{tab:maxpixels}
\begin{tabular}{lccccc}
\multirow{2}{*}{\textbf{Max Pixels}} & \multirow{2}{*}{\textbf{Resolution}} & \multicolumn{2}{c}{\textbf{EventClassification}} & \multicolumn{2}{c}{\textbf{VideoQA}} \\
 &  & \textbf{0} & \textbf{1} & \textbf{0} & \textbf{1} \\
\hline
10,000 & 128 $\times$ 72  & 1146 & 314 & 1127 & 333 \\

40,000 & 256 $\times$ 144 & 797 & 663 & 976 & 483 \\

\hl{\textbf{100,352}} & 416 $\times$ 234 & 567 & 893 & 814 & 646 \\

160,000 & 528 $\times$ 297 & 535 & 925 & 786 & 674 \\

250,000 & 656 $\times$ 369 & 519 & 941 & 763 & 697 \\
\hline
\end{tabular}
\end{table}

The corresponding ROC-AUC values in Tables~\ref{tab:frame_performance} and~\ref{tab:maxpixels_performance} reveal only modest and somewhat irregular dependencies on spatiotemporal settings. For frame counts, VASE on EventClassification ranges between 0.526 and 0.654, with relatively high values already at 4 frames (0.626), a notable dip at 8 frames (0.526), and local peaks at 16 and 24 frames (0.652 and 0.654). On VideoQA, VASE increases from 0.552 (8 frames) to 0.649 (20 frames), but then slightly decreases at 24 and 30 frames (0.623 and 0.612). These patterns suggest that providing more temporal evidence than the very low 4 to 8 frame settings generally helps, but benefits beyond roughly 16 to 24 frames are small and not monotonic.

\begin{table}[htb]
\centering
\small
\caption{ROC-AUC performance for EventClassification and VideoQA across the number of frames sampled per video using embedding-based clustering with SoccerChat-qwen2-vl.}
\label{tab:frame_performance}
\begin{tabular}{lcccccc}
\textbf{Frames} & \multicolumn{3}{c}{\textbf{EventClassification}} & \multicolumn{3}{c}{\textbf{VideoQA}} \\
 & \textbf{SE} & \textbf{RadFlag} & \textbf{VASE} & \textbf{SE} & \textbf{RadFlag} & \textbf{VASE} \\
\hline
4  & 0.538 & 0.538 & 0.626 & 0.537 & 0.530 & 0.608 \\
8  & 0.488 & 0.498 & 0.526 & 0.518 & 0.514 & 0.552 \\
12 & 0.507 & 0.512 & 0.614 & 0.540 & 0.535 & 0.636 \\
16 & 0.543 & 0.530 & 0.652 & 0.535 & 0.536 & 0.637 \\
20 & 0.505 & 0.516 & 0.618 & 0.555 & 0.550 & \hl{\textbf{0.649}} \\
\hl{\textbf{24}} & 0.513 & 0.516 & \hl{\textbf{0.654}} & 0.524 & 0.523 & 0.623 \\
30 & 0.517 & 0.509 & 0.602 & 0.547 & 0.541 & 0.612 \\
\hline
\end{tabular}
\end{table}

\begin{table}[htb]
\centering
\small
\caption{ROC-AUC performance for EventClassification and VideoQA across maximum pixel constraints using embedding-based clustering with SoccerChat-qwen2-vl.}
\label{tab:maxpixels_performance}
\begin{tabular}{lcccccc}
\textbf{Max Pixels} & \multicolumn{3}{c}{\textbf{EventClassification}} & \multicolumn{3}{c}{\textbf{VideoQA}} \\
 & \textbf{SE} & \textbf{RadFlag} & \textbf{VASE} & \textbf{SE} & \textbf{RadFlag} & \textbf{VASE} \\
\hline
10,000   & 0.531 & 0.548 & 0.573 & 0.505 & 0.521 & 0.576 \\
40,000   & 0.534 & 0.526 & 0.600 & 0.545 & 0.544 & 0.621 \\
\hl{\textbf{100,352}} & 0.513 & 0.516 & \hl{\textbf{0.654}} & 0.524 & 0.523 & 0.623 \\
160,000  & 0.569 & 0.573 & 0.630 & 0.546 & 0.568 & 0.628 \\
250,000  & 0.530 & 0.532 & 0.648 & 0.561 & 0.557 & \hl{\textbf{0.647}} \\
\hline

\end{tabular}
\end{table}

For spatial resolution, VASE on EventClassification increases from 0.573 at 10{,}000 pixels to 0.654 at 100{,}352 pixels, then slightly decreases to 0.630 at 160{,}000 pixels. On VideoQA, VASE steadily improves from 0.576 (10{,}000) through 0.621 (40{,}000) and 0.623 (100{,}352) to 0.628 (160{,}000), with the highest value (0.647) appearing at 250{,}000 pixels. Taken together, these results support a simple picture: extremely low resolutions degrade both accuracy and detectability of hallucinations, while moderate to high resolutions improve both, with only mild and somewhat noisy gains beyond the mid-range.

\textbf{Model comparison: architecture vs.\ domain adaptation.}
Finally, we relate these metric behaviours back to model design. SoccerChat-qwen2-vl, which is fine-tuned on the SoccerChat dataset, produces substantially fewer hallucinations than either Qwen2-VL or Qwen2.5-VL. However, the corresponding VASE AUC values in Table~\ref{tab:hedge_video_auc} show only moderate improvements in calibration: for EventClassification under embedding-based clustering, VASE is 0.654 for SoccerChat versus 0.644 for Qwen2.5-VL; for VideoQA, SoccerChat reaches 0.623--0.631, comparable to Qwen2.5-VL’s best 0.640. This suggests that domain adaptation primarily reduces the frequency of hallucinations, while improvements in the ability of reliability metrics to separate correct from incorrect answers are more modest. Across all three models, the qualitative pattern is similar: SE and RadFlag sometimes help but can be close to chance, whereas VASE, especially at larger distortion budgets, tends to provide the most informative signal.

Overall, the results indicate that perturbation-based, cluster-level metrics such as VASE offer a meaningful but still imperfect handle on hallucination detection in VideoVQA. They outperform clean-only uncertainty estimates in many settings and scale well with embedding-based clustering, yet absolute ROC-AUC remains moderate, and several trends with respect to frames and resolution are noisy. Addressing these limitations by expanding the completed experimental configurations, including human adjudication on a subset of examples, and exploring richer perturbation families remain open directions for future work.

\section{Discussion}
\label{sec:discussion}

The empirical findings above reveal several properties of hallucinations in contemporary Video-VLMs and clarify where VideoHEDGE helps. Our first observation concerns the limits of unimodal uncertainty. For weaker architectures such as Qwen2-VL on EventClassification, SE and RadFlag operate close to chance, even though hallucinations are frequent. This pattern is consistent with hallucinations that are driven by high-confidence language priors: the model repeatedly produces the same incorrect answer with a sharply peaked semantic distribution over clean samples, so dispersion-based signals fail to indicate a lack of grounding. At the same time, stronger models such as Qwen2.5-VL achieve SE and RadFlag values around 0.6 in some settings, indicating that architectural improvements can make clean-only uncertainty partially informative, but not sufficient on their own.

VASE is designed precisely to probe robustness rather than static confidence, by measuring the divergence between semantic distributions induced by clean and perturbed visual inputs. Across models, tasks, and clustering backends, it typically provides the strongest or near-strongest separation between hallucinated and supported answers. The distortion-budget study shows that this advantage depends on having enough visual variation: with only one or two distortions, VASE is only mildly better than chance, but as we move into the 6--10 distortion regime, AUC improves substantially, while SE and RadFlag remain flat. This suggests that single perturbations are often too weak to expose unstable visual grounding, and that composite distortions combining brightness, contrast, saturation, hue changes, and noise are needed to reliably stress-test predictions. The gains at very high budgets are not strictly monotonic and appear to taper off, but the best-performing configurations consistently occur with larger distortion sets, indicating diminishing but still positive returns.

Our second observation concerns the distinction between task accuracy and calibration. Fine-tuning on SoccerChat dramatically reduces the number of hallucinations relative to generalist models, yet the corresponding VASE scores improve only modestly and are similar to those of Qwen2.5-VL. This indicates that standard domain fine-tuning sharpens the model's factual knowledge about soccer, but does not automatically align its confidence with correctness: when the fine-tuned model is wrong, its internal uncertainty still looks similar to that of generalist models. Closing this gap likely requires reliability-aware training objectives that explicitly shape uncertainty (e.g., by rewarding calibrated confidence or penalizing overconfident errors), which our current experiments do not attempt but naturally point to as future work.

The spatiotemporal sensitivity analysis offers a coarse picture of how much visual evidence is needed for meaningful reliability estimation. Extremely low resolutions clearly degrade both accuracy and detection performance. For frame counts, we observe a notable dip in VASE at 8 frames, whereas 4-frame and 12--24-frame settings achieve comparable or higher AUCs. Once a moderate number of frames (roughly 12--24) and pixels (40{,}000--160{,}000 per frame) are available, VASE AUC is generally higher than in the worst 8-frame setting, but further increases yield only small and somewhat noisy gains, while correctness remains relatively stable across frame counts. The slight drop in performance observed at 30 frames suggests that simply adding more temporal context does not necessarily help and may complicate the model's attention patterns, though we do not directly measure attention and therefore treat this as a plausible, not confirmed, explanation. Overall, the results support using reasonably informative but not extreme spatiotemporal settings when deploying perturbation-based reliability metrics.

Finally, the comparison between clustering backends challenges the practical role of NLI as a default ``gold standard'' for semantic clustering in this context. While NLI-based clustering provides explicit entailment and contradiction judgements, its empirical advantage in ROC--AUC is small and inconsistent: in some configurations it slightly outperforms embeddings, in others it slightly underperforms. In contrast, the runtime gap grows quickly with the number of samples, making NLI-based clustering prohibitively expensive at larger distortion budgets. One possible explanation is that NLI models are sensitive to minor phrasings in short captions, whereas sentence embeddings capture a more forgiving ``semantic gist'' that is sufficient for clustering answers into hypotheses at the granularity required by VideoHEDGE. Regardless of the exact cause, our measurements indicate that embedding-based clustering achieves nearly the same detection performance at orders-of-magnitude lower cost, and is therefore the only practical choice for large-scale video reliability analysis.

\section{Conclusion}
\label{sec:conclusion}

VideoHEDGE extends the HEDGE framework to video by combining spatiotemporal sampling, semantic clustering, and perturbation-aware scoring for hallucination detection in Video-VLMs. On the SoccerChat benchmark, our experiments show that standard clean-only uncertainty metrics such as Semantic Entropy (SE) and RadFlag can be close to random in challenging settings, particularly for weaker architectures, while Vision-Amplified Semantic Entropy (VASE) consistently provides stronger separation between supported and hallucinated answers. This improvement is most pronounced when the metric is given a sufficiently rich set of visual perturbations, confirming the value of treating reliability as robustness under controlled degradations.

We also find that domain adaptation and architecture affect different aspects of reliability. Fine-tuning on SoccerChat substantially reduces hallucination rates but yields only modest gains in ROC--AUC, whereas architectural improvements from Qwen2-VL to Qwen2.5-VL noticeably strengthen clean-only uncertainty metrics in some cases. These results underscore that accuracy and calibration are partially decoupled: a model can become more often correct without becoming much easier to diagnose when it is wrong.

From a system perspective, our comparison of clustering backends shows that embedding-based clustering offers nearly the same detection performance as NLI-based clustering, while scaling far better with the number of samples. This makes embeddings the practical backbone for VideoHEDGE, with NLI serving as an expensive reference rather than a default choice. Finally, our spatiotemporal and resolution sweeps suggest that reliability estimation benefits from moderate visual fidelity but exhibits diminishing returns and some instability at very high frame counts and resolutions.

Overall, VideoHEDGE provides a scalable, visually grounded framework that turns hallucination detection for Video-VLMs into a moderately informative diagnostic rather than a near-random guess. At the same time, the absolute AUC values, noisy trends across settings, and reliance on an LLM adjudicator highlight that robust, well-calibrated reliability estimation for video remains an open challenge. Potential future work includes integrating reliability-aware objectives into training, broadening the family of perturbations, and incorporating human adjudication to further validate and refine perturbation-based reliability metrics.

\begin{acks}
This work was partly funded by the Research Council of Norway, project number 346671 (AI-Storyteller), and has benefited from the Experimental Infrastructure for Exploration of Exascale Computing (eX3), which is financially supported by the Research Council of Norway under contract 270053.

\end{acks}

\bibliographystyle{ACM-Reference-Format}
\bibliography{references}


\begin{thebibliography}{43}


\ifx \showCODEN    \undefined \def \showCODEN     #1{\unskip}     \fi
\ifx \showISBNx    \undefined \def \showISBNx     #1{\unskip}     \fi
\ifx \showISBNxiii \undefined \def \showISBNxiii  #1{\unskip}     \fi
\ifx \showISSN     \undefined \def \showISSN      #1{\unskip}     \fi
\ifx \showLCCN     \undefined \def \showLCCN      #1{\unskip}     \fi
\ifx \shownote     \undefined \def \shownote      #1{#1}          \fi
\ifx \showarticletitle \undefined \def \showarticletitle #1{#1}   \fi
\ifx \showURL      \undefined \def \showURL       {\relax}        \fi
\providecommand\bibfield[2]{#2}
\providecommand\bibinfo[2]{#2}
\providecommand\natexlab[1]{#1}
\providecommand\showeprint[2][]{arXiv:#2}

\bibitem[Abdaljalil et~al\mbox{.}(2025)]%
        {Abdaljalil2025Mar}
\bibfield{author}{\bibinfo{person}{Samir Abdaljalil}, \bibinfo{person}{Hasan Kurban}, \bibinfo{person}{Parichit Sharma}, {et~al\mbox{.}}} \bibinfo{year}{2025}\natexlab{}.
\newblock \showarticletitle{{SINdex: Semantic INconsistency Index for Hallucination Detection in LLMs}}.
\newblock \bibinfo{journal}{\emph{ArXiv e-prints}} (\bibinfo{date}{March} \bibinfo{year}{2025}).
\newblock
\href{https://doi.org/10.48550/arXiv.2503.05980}{doi:\nolinkurl{10.48550/arXiv.2503.05980}}


\bibitem[Bai et~al\mbox{.}(2025)]%
        {Qwen25vl}
\bibfield{author}{\bibinfo{person}{Shuai Bai}, \bibinfo{person}{Keqin Chen}, \bibinfo{person}{Xuejing Liu}, {et~al\mbox{.}}} \bibinfo{year}{2025}\natexlab{}.
\newblock \showarticletitle{{Qwen2.5-VL Technical Report}}.
\newblock \bibinfo{journal}{\emph{ArXiv e-prints}} (\bibinfo{date}{Feb.} \bibinfo{year}{2025}).
\newblock
\href{https://doi.org/10.48550/arXiv.2502.13923}{doi:\nolinkurl{10.48550/arXiv.2502.13923}}


\bibitem[Chen et~al\mbox{.}(2024)]%
        {mo_hallucination_2024}
\bibfield{author}{\bibinfo{person}{Xuweiyi Chen}, \bibinfo{person}{Ziqiao Ma}, \bibinfo{person}{Xuejun Zhang}, {et~al\mbox{.}}} \bibinfo{year}{2024}\natexlab{}.
\newblock \showarticletitle{{Multi-Object Hallucination in Vision-Language Models}}.
\newblock \bibinfo{journal}{\emph{ArXiv e-prints}} (\bibinfo{date}{July} \bibinfo{year}{2024}).
\newblock
\href{https://doi.org/10.48550/arXiv.2407.06192}{doi:\nolinkurl{10.48550/arXiv.2407.06192}}


\bibitem[Chen et~al\mbox{.}(2025)]%
        {MMSurvey2025}
\bibfield{author}{\bibinfo{person}{Zhiyuan Chen}, \bibinfo{person}{Yuecong Min}, \bibinfo{person}{Jie Zhang}, {et~al\mbox{.}}} \bibinfo{year}{2025}\natexlab{}.
\newblock \showarticletitle{{A Survey of Multimodal Hallucination Evaluation and Detection}}.
\newblock \bibinfo{journal}{\emph{ArXiv e-prints}} (\bibinfo{date}{July} \bibinfo{year}{2025}).
\newblock
\href{https://doi.org/10.48550/arXiv.2507.19024}{doi:\nolinkurl{10.48550/arXiv.2507.19024}}


\bibitem[Deli{\ifmmode\grave{e}\else\`{e}\fi}ge et~al\mbox{.}(2021)]%
        {soccernet_v2}
\bibfield{author}{\bibinfo{person}{Adrien Deli{\ifmmode\grave{e}\else\`{e}\fi}ge}, \bibinfo{person}{Anthony Cioppa}, \bibinfo{person}{Silvio Giancola}, {et~al\mbox{.}}} \bibinfo{year}{2021}\natexlab{}.
\newblock \bibinfo{booktitle}{\emph{{SoccerNet-v2: A Dataset and Benchmarks for Holistic Understanding of Broadcast Soccer Videos}}}.
\newblock \bibinfo{publisher}{IEEE Computer Society}.
\newblock
\showISBNx{978-1-6654-4899-4}
\href{https://doi.org/10.1109/CVPRW53098.2021.00508}{doi:\nolinkurl{10.1109/CVPRW53098.2021.00508}}


\bibitem[Farquhar et~al\mbox{.}(2024)]%
        {SE2024}
\bibfield{author}{\bibinfo{person}{Sebastian Farquhar}, \bibinfo{person}{Jannik Kossen}, \bibinfo{person}{Lorenz Kuhn}, {and} \bibinfo{person}{Yarin Gal}.} \bibinfo{year}{2024}\natexlab{}.
\newblock \showarticletitle{{Detecting hallucinations in large language models using semantic entropy}}.
\newblock \bibinfo{journal}{\emph{Nature}} \bibinfo{volume}{630}, \bibinfo{number}{8017} (\bibinfo{date}{June} \bibinfo{year}{2024}), \bibinfo{pages}{625--630}.
\newblock
\showISSN{1476-4687}
\href{https://doi.org/10.1038/s41586-024-07421-0}{doi:\nolinkurl{10.1038/s41586-024-07421-0}}


\bibitem[Fawcett(2006)]%
        {fawcett2006ROC}
\bibfield{author}{\bibinfo{person}{Tom Fawcett}.} \bibinfo{year}{2006}\natexlab{}.
\newblock \showarticletitle{An introduction to ROC analysis}.
\newblock \bibinfo{journal}{\emph{Pattern recognition letters}} \bibinfo{volume}{27}, \bibinfo{number}{8} (\bibinfo{year}{2006}), \bibinfo{pages}{861--874}.
\newblock


\bibitem[{FFmpeg Developers}(2025)]%
        {ffmpeg}
\bibfield{author}{\bibinfo{person}{{FFmpeg Developers}}.} \bibinfo{year}{2025}\natexlab{}.
\newblock \bibinfo{title}{FFmpeg}.
\newblock
\urldef\tempurl%
\url{https://ffmpeg.org/}
\showURL{%
\tempurl}


\bibitem[Gao et~al\mbox{.}(2024)]%
        {gao2024spuq}
\bibfield{author}{\bibinfo{person}{Xiang Gao}, \bibinfo{person}{Jiaxin Zhang}, \bibinfo{person}{Lalla Mouatadid}, {and} \bibinfo{person}{Kamalika Das}.} \bibinfo{year}{2024}\natexlab{}.
\newblock \showarticletitle{SPUQ: Perturbation-Based Uncertainty Quantification for Large Language Models}. In \bibinfo{booktitle}{\emph{Proceedings of EACL (Long Papers)}}. \bibinfo{pages}{2336--2346}.
\newblock
\href{https://doi.org/10.18653/v1/2024.eacl-long.143}{doi:\nolinkurl{10.18653/v1/2024.eacl-long.143}}


\bibitem[Gautam et~al\mbox{.}(2025a)]%
        {gautam2025soccerchat}
\bibfield{author}{\bibinfo{person}{Sushant Gautam}, \bibinfo{person}{Cise Midoglu}, \bibinfo{person}{Vajira~Lasantha Thambawita}, \bibinfo{person}{Michael~A. Riegler}, \bibinfo{person}{P{\aa}l Halvorsen}, {and} \bibinfo{person}{Mubarak Shah}.} \bibinfo{year}{2025}\natexlab{a}.
\newblock \showarticletitle{SoccerChat: Integrating Multimodal Data for Enhanced Soccer Game Understanding}. In \bibinfo{booktitle}{\emph{Proceedings of the IEEE International Conference on Content-Based Multimedia Indexing (CBMI)}}. \bibinfo{address}{IEEE CBMI 2025}.
\newblock


\bibitem[Gautam et~al\mbox{.}(2025b)]%
        {HEDGE}
\bibfield{author}{\bibinfo{person}{Sushant Gautam}, \bibinfo{person}{Michael~A. Riegler}, {and} \bibinfo{person}{P{\aa}l Halvorsen}.} \bibinfo{year}{2025}\natexlab{b}.
\newblock \showarticletitle{{HEDGE: Hallucination Estimation via Dense Geometric Entropy for VQA with Vision-Language Models}}.
\newblock \bibinfo{journal}{\emph{ArXiv e-prints}} (\bibinfo{date}{Nov.} \bibinfo{year}{2025}).
\newblock
\href{https://doi.org/10.48550/arXiv.2511.12693}{doi:\nolinkurl{10.48550/arXiv.2511.12693}}


\bibitem[Geng et~al\mbox{.}(2023)]%
        {calibration_llms_survey_2024}
\bibfield{author}{\bibinfo{person}{Jiahui Geng}, \bibinfo{person}{Fengyu Cai}, \bibinfo{person}{Yuxia Wang}, {et~al\mbox{.}}} \bibinfo{year}{2023}\natexlab{}.
\newblock \showarticletitle{{A Survey of Confidence Estimation and Calibration in Large Language Models}}.
\newblock \bibinfo{journal}{\emph{ArXiv e-prints}} (\bibinfo{date}{Nov.} \bibinfo{year}{2023}).
\newblock
\href{https://doi.org/10.48550/arXiv.2311.08298}{doi:\nolinkurl{10.48550/arXiv.2311.08298}}


\bibitem[Giancola et~al\mbox{.}(2018)]%
        {soccernet_v1}
\bibfield{author}{\bibinfo{person}{Silvio Giancola}, \bibinfo{person}{Mohieddine Amine}, \bibinfo{person}{Tarek Dghaily}, {et~al\mbox{.}}} \bibinfo{year}{2018}\natexlab{}.
\newblock \bibinfo{booktitle}{\emph{{SoccerNet: A Scalable Dataset for Action Spotting in Soccer Videos}}}.
\newblock \bibinfo{publisher}{IEEE Computer Society}.
\newblock
\showISBNx{978-1-5386-6100-0}
\href{https://doi.org/10.1109/CVPRW.2018.00223}{doi:\nolinkurl{10.1109/CVPRW.2018.00223}}


\bibitem[Giancola et~al\mbox{.}(2022)]%
        {Giancola2022Oct}
\bibfield{author}{\bibinfo{person}{Silvio Giancola}, \bibinfo{person}{Anthony Cioppa}, \bibinfo{person}{Adrien Deli{\ifmmode\grave{e}\else\`{e}\fi}ge}, {et~al\mbox{.}}} \bibinfo{year}{2022}\natexlab{}.
\newblock \showarticletitle{{SoccerNet 2022 Challenges Results}}.
\newblock In \bibinfo{booktitle}{\emph{{ACM Conferences}}}. \bibinfo{publisher}{Association for Computing Machinery}, \bibinfo{address}{New York, NY, USA}, \bibinfo{pages}{75--86}.
\newblock
\href{https://doi.org/10.1145/3552437.3558545}{doi:\nolinkurl{10.1145/3552437.3558545}}


\bibitem[He et~al\mbox{.}(2020)]%
        {DeBERTaHe2020Jun}
\bibfield{author}{\bibinfo{person}{Pengcheng He}, \bibinfo{person}{Xiaodong Liu}, \bibinfo{person}{Jianfeng Gao}, {et~al\mbox{.}}} \bibinfo{year}{2020}\natexlab{}.
\newblock \showarticletitle{{DeBERTa: Decoding-enhanced BERT with Disentangled Attention}}.
\newblock \bibinfo{journal}{\emph{ArXiv e-prints}} (\bibinfo{date}{June} \bibinfo{year}{2020}).
\newblock
\href{https://doi.org/10.48550/arXiv.2006.03654}{doi:\nolinkurl{10.48550/arXiv.2006.03654}}


\bibitem[Held et~al\mbox{.}(2024)]%
        {xvars_2024}
\bibfield{author}{\bibinfo{person}{Jan Held}, \bibinfo{person}{Hani Itani}, \bibinfo{person}{Anthony Cioppa}, {et~al\mbox{.}}} \bibinfo{year}{2024}\natexlab{}.
\newblock \bibinfo{booktitle}{\emph{{X-VARS: Introducing Explainability in Football Refereeing with Multi-Modal Large Language Models}}}.
\newblock \bibinfo{publisher}{IEEE Computer Society}.
\newblock
\showISBNx{979-8-3503-6547-4}
\href{https://doi.org/10.1109/CVPRW63382.2024.00332}{doi:\nolinkurl{10.1109/CVPRW63382.2024.00332}}


\bibitem[Hu et~al\mbox{.}(2024)]%
        {survey_vidllm_2024}
\bibfield{author}{\bibinfo{person}{Zi-Yuan Hu}, \bibinfo{person}{Yiwu Zhong}, \bibinfo{person}{Shijia Huang}, {et~al\mbox{.}}} \bibinfo{year}{2024}\natexlab{}.
\newblock \showarticletitle{{Enhancing Temporal Modeling of Video LLMs via Time Gating}}.
\newblock \bibinfo{journal}{\emph{ACL Anthology}} (\bibinfo{date}{Nov.} \bibinfo{year}{2024}), \bibinfo{pages}{2845--2856}.
\newblock
\href{https://doi.org/10.18653/v1/2024.findings-emnlp.162}{doi:\nolinkurl{10.18653/v1/2024.findings-emnlp.162}}


\bibitem[Huang et~al\mbox{.}(2025)]%
        {ACMSurvey2025}
\bibfield{author}{\bibinfo{person}{Lei Huang}, \bibinfo{person}{Weijiang Yu}, \bibinfo{person}{Weitao Ma}, {et~al\mbox{.}}} \bibinfo{year}{2025}\natexlab{}.
\newblock \showarticletitle{{A Survey on Hallucination in Large Language Models: Principles, Taxonomy, Challenges, and Open Questions}}.
\newblock \bibinfo{journal}{\emph{ACM Transactions on Information Systems}} \bibinfo{volume}{43}, \bibinfo{number}{2} (\bibinfo{date}{Jan.} \bibinfo{year}{2025}), \bibinfo{pages}{1--55}.
\newblock
\showISSN{1046-8188}
\href{https://doi.org/10.1145/3703155}{doi:\nolinkurl{10.1145/3703155}}


\bibitem[Kaul et~al\mbox{.}(2024)]%
        {THRONE2024}
\bibfield{author}{\bibinfo{person}{Prannay Kaul}, \bibinfo{person}{Zhizhong Li}, \bibinfo{person}{Hao Yang}, {et~al\mbox{.}}} \bibinfo{year}{2024}\natexlab{}.
\newblock \showarticletitle{{THRONE: An Object-based Hallucination Benchmark for the Free-form Generations of Large Vision-Language Models}}.
\newblock \bibinfo{journal}{\emph{ArXiv e-prints}} (\bibinfo{date}{May} \bibinfo{year}{2024}).
\newblock
\href{https://doi.org/10.48550/arXiv.2405.05256}{doi:\nolinkurl{10.48550/arXiv.2405.05256}}


\bibitem[Kossen et~al\mbox{.}(2024)]%
        {Kossen2024SEP}
\bibfield{author}{\bibinfo{person}{Jannik Kossen}, \bibinfo{person}{Jiatong Han}, \bibinfo{person}{Muhammed Razzak}, {et~al\mbox{.}}} \bibinfo{year}{2024}\natexlab{}.
\newblock \showarticletitle{{Semantic Entropy Probes: Robust and Cheap Hallucination Detection in LLMs}}.
\newblock \bibinfo{journal}{\emph{ArXiv e-prints}} (\bibinfo{date}{June} \bibinfo{year}{2024}).
\newblock
\href{https://doi.org/10.48550/arXiv.2406.15927}{doi:\nolinkurl{10.48550/arXiv.2406.15927}}


\bibitem[Kostumov et~al\mbox{.}(2024)]%
        {llm_vlm_calibration_2024}
\bibfield{author}{\bibinfo{person}{Vasily Kostumov}, \bibinfo{person}{Bulat Nutfullin}, \bibinfo{person}{Oleg Pilipenko}, {et~al\mbox{.}}} \bibinfo{year}{2024}\natexlab{}.
\newblock \showarticletitle{{Uncertainty-Aware Evaluation for Vision-Language Models}}.
\newblock \bibinfo{journal}{\emph{ArXiv e-prints}} (\bibinfo{date}{Feb.} \bibinfo{year}{2024}).
\newblock
\href{https://doi.org/10.48550/arXiv.2402.14418}{doi:\nolinkurl{10.48550/arXiv.2402.14418}}


\bibitem[Lan et~al\mbox{.}(2024)]%
        {survey_lvml_hallucination_lan_2024}
\bibfield{author}{\bibinfo{person}{Wei Lan}, \bibinfo{person}{Wenyi Chen}, \bibinfo{person}{Qingfeng Chen}, {et~al\mbox{.}}} \bibinfo{year}{2024}\natexlab{}.
\newblock \showarticletitle{{A Survey of Hallucination in Large Visual Language Models}}.
\newblock \bibinfo{journal}{\emph{ArXiv e-prints}} (\bibinfo{date}{Oct.} \bibinfo{year}{2024}).
\newblock
\href{https://doi.org/10.48550/arXiv.2410.15359}{doi:\nolinkurl{10.48550/arXiv.2410.15359}}


\bibitem[Lau et~al\mbox{.}(2025)]%
        {umpire_2025}
\bibfield{author}{\bibinfo{person}{Gregory Kang~Ruey Lau}, \bibinfo{person}{Hieu Dao}, \bibinfo{person}{Nicole Kan~Hui Lin}, {and} \bibinfo{person}{Bryan Kian~Hsiang Low}.} \bibinfo{year}{2025}\natexlab{}.
\newblock \showarticletitle{Uncertainty Quantification for Multimodal Large Language Models}. In \bibinfo{booktitle}{\emph{Proceedings of the ICML 2025 Workshop on Reliable and Responsible Foundation Models (R2-FM)}}.
\newblock


\bibitem[Lee et~al\mbox{.}(2024)]%
        {vcd_2024}
\bibfield{author}{\bibinfo{person}{Yi-Lun Lee}, \bibinfo{person}{Yi-Hsuan Tsai}, {and} \bibinfo{person}{Wei-Chen Chiu}.} \bibinfo{year}{2024}\natexlab{}.
\newblock \showarticletitle{{Delve into Visual Contrastive Decoding for Hallucination Mitigation of Large Vision-Language Models}}.
\newblock \bibinfo{journal}{\emph{ArXiv e-prints}} (\bibinfo{date}{Dec.} \bibinfo{year}{2024}).
\newblock
\href{https://doi.org/10.48550/arXiv.2412.06775}{doi:\nolinkurl{10.48550/arXiv.2412.06775}}


\bibitem[Liao et~al\mbox{.}(2025)]%
        {VASE2025}
\bibfield{author}{\bibinfo{person}{Zehui Liao}, \bibinfo{person}{Shishuai Hu}, \bibinfo{person}{Ke Zou}, \bibinfo{person}{Huazhu Fu}, \bibinfo{person}{Liangli Zhen}, {et~al\mbox{.}}} \bibinfo{year}{2025}\natexlab{}.
\newblock \showarticletitle{{Vision-Amplified Semantic Entropy for Hallucination Detection in Medical Visual Question Answering}}.
\newblock \bibinfo{journal}{\emph{ArXiv e-prints}} (\bibinfo{date}{March} \bibinfo{year}{2025}).
\newblock
\href{https://doi.org/10.48550/arXiv.2503.20504}{doi:\nolinkurl{10.48550/arXiv.2503.20504}}


\bibitem[Liu et~al\mbox{.}(2024)]%
        {survey_lvml_hallucination_liu_2024}
\bibfield{author}{\bibinfo{person}{Hanchao Liu}, \bibinfo{person}{Wenyuan Xue}, \bibinfo{person}{Yifei Chen}, {et~al\mbox{.}}} \bibinfo{year}{2024}\natexlab{}.
\newblock \showarticletitle{{A Survey on Hallucination in Large Vision-Language Models}}.
\newblock \bibinfo{journal}{\emph{ArXiv e-prints}} (\bibinfo{date}{Feb.} \bibinfo{year}{2024}).
\newblock
\href{https://doi.org/10.48550/arXiv.2402.00253}{doi:\nolinkurl{10.48550/arXiv.2402.00253}}


\bibitem[Liu et~al\mbox{.}(2025)]%
        {uq_llms_survey_2025}
\bibfield{author}{\bibinfo{person}{Xiaoou Liu}, \bibinfo{person}{Tiejin Chen}, \bibinfo{person}{Longchao Da}, {et~al\mbox{.}}} \bibinfo{year}{2025}\natexlab{}.
\newblock \showarticletitle{{Uncertainty Quantification and Confidence Calibration in Large Language Models: A Survey}}.
\newblock \bibinfo{journal}{\emph{ArXiv e-prints}} (\bibinfo{date}{March} \bibinfo{year}{2025}).
\newblock
\href{https://doi.org/10.48550/arXiv.2503.15850}{doi:\nolinkurl{10.48550/arXiv.2503.15850}}


\bibitem[P.~J. and Kovoor(2024)]%
        {survey_videoqa_2024}
\bibfield{author}{\bibinfo{person}{Jeshmol P.~J.} {and} \bibinfo{person}{Binsu~C. Kovoor}.} \bibinfo{year}{2024}\natexlab{}.
\newblock \showarticletitle{{Video Question Answering: A survey of the state-of-the-art}}.
\newblock \bibinfo{journal}{\emph{Journal of Visual Communication and Image Representation}}  \bibinfo{volume}{105} (\bibinfo{date}{Dec.} \bibinfo{year}{2024}), \bibinfo{pages}{104320}.
\newblock
\showISSN{1047-3203}
\href{https://doi.org/10.1016/j.jvcir.2024.104320}{doi:\nolinkurl{10.1016/j.jvcir.2024.104320}}


\bibitem[Padhi et~al\mbox{.}(2025)]%
        {padhi2025multimodal}
\bibfield{author}{\bibinfo{person}{Trilok Padhi}, \bibinfo{person}{Ramneet Kaur}, \bibinfo{person}{Adam~D. Cobb}, {et~al\mbox{.}}} \bibinfo{year}{2025}\natexlab{}.
\newblock \showarticletitle{{Calibrating Uncertainty Quantification of Multi-Modal LLMs using Grounding}}.
\newblock \bibinfo{journal}{\emph{ArXiv e-prints}} (\bibinfo{date}{April} \bibinfo{year}{2025}).
\newblock
\href{https://doi.org/10.48550/arXiv.2505.03788}{doi:\nolinkurl{10.48550/arXiv.2505.03788}}


\bibitem[Qi et~al\mbox{.}(2024)]%
        {AHESurvey2024}
\bibfield{author}{\bibinfo{person}{Siya Qi}, \bibinfo{person}{Lin Gui}, \bibinfo{person}{Yulan He}, {et~al\mbox{.}}} \bibinfo{year}{2024}\natexlab{}.
\newblock \showarticletitle{{A Survey of Automatic Hallucination Evaluation on Natural Language Generation}}.
\newblock \bibinfo{journal}{\emph{ArXiv e-prints}} (\bibinfo{date}{April} \bibinfo{year}{2024}).
\newblock
\href{https://doi.org/10.48550/arXiv.2404.12041}{doi:\nolinkurl{10.48550/arXiv.2404.12041}}


\bibitem[Quan et~al\mbox{.}(2025)]%
        {contrastive_retrieval_2024}
\bibfield{author}{\bibinfo{person}{Guofeng Quan}, \bibinfo{person}{Wenfeng Feng}, \bibinfo{person}{Chuzhan Hao}, \bibinfo{person}{Guochao Jiang}, \bibinfo{person}{Yuewei Zhang}, {and} \bibinfo{person}{Hao~Henry Wang}.} \bibinfo{year}{2025}\natexlab{}.
\newblock \showarticletitle{{RASD}: Retrieval-Augmented Speculative Decoding}. In \bibinfo{booktitle}{\emph{Findings of the Association for Computational Linguistics: ACL 2025}}, \bibfield{editor}{\bibinfo{person}{Wanxiang Che}, \bibinfo{person}{Joyce Nabende}, \bibinfo{person}{Ekaterina Shutova}, {and} \bibinfo{person}{Mohammad~Taher Pilehvar}} (Eds.). \bibinfo{publisher}{Association for Computational Linguistics}, \bibinfo{address}{Vienna, Austria}, \bibinfo{pages}{6167--6177}.
\newblock
\showISBNx{979-8-89176-256-5}
\href{https://doi.org/10.18653/v1/2025.findings-acl.320}{doi:\nolinkurl{10.18653/v1/2025.findings-acl.320}}


\bibitem[Reimers and Gurevych(2019)]%
        {reimers2019sentence}
\bibfield{author}{\bibinfo{person}{Nils Reimers} {and} \bibinfo{person}{Iryna Gurevych}.} \bibinfo{year}{2019}\natexlab{}.
\newblock \showarticletitle{Sentence-BERT: Sentence Embeddings using Siamese BERT Networks}. In \bibinfo{booktitle}{\emph{Proceedings of the 2019 Conference on Empirical Methods in Natural Language Processing and the 9th International Joint Conference on Natural Language Processing (EMNLP–IJCNLP)}}. Association for Computational Linguistics, \bibinfo{address}{Hong Kong, China}, \bibinfo{pages}{3982--3992}.
\newblock
\href{https://doi.org/10.18653/v1/D19-1410}{doi:\nolinkurl{10.18653/v1/D19-1410}}


\bibitem[Rohrbach et~al\mbox{.}(2018)]%
        {ObjectHalBench2024}
\bibfield{author}{\bibinfo{person}{Anna Rohrbach}, \bibinfo{person}{Lisa~Anne Hendricks}, \bibinfo{person}{Kaylee Burns}, {et~al\mbox{.}}} \bibinfo{year}{2018}\natexlab{}.
\newblock \showarticletitle{{Object Hallucination in Image Captioning}}.
\newblock \bibinfo{journal}{\emph{ACL Anthology}} (\bibinfo{year}{2018}), \bibinfo{pages}{4035--4045}.
\newblock
\href{https://doi.org/10.18653/v1/D18-1437}{doi:\nolinkurl{10.18653/v1/D18-1437}}


\bibitem[Sahoo et~al\mbox{.}(2024)]%
        {survey_lvml_hallucination_sahoo_2024}
\bibfield{author}{\bibinfo{person}{Pranab Sahoo}, \bibinfo{person}{Prabhash Meharia}, \bibinfo{person}{Akash Ghosh}, \bibinfo{person}{Sriparna Saha}, \bibinfo{person}{Vinija Jain}, {and} \bibinfo{person}{Aman Chadha}.} \bibinfo{year}{2024}\natexlab{}.
\newblock \showarticletitle{A Comprehensive Survey of Hallucination in Large Language, Image, Video and Audio Foundation Models}. In \bibinfo{booktitle}{\emph{Findings of the Association for Computational Linguistics: EMNLP 2024}}, \bibfield{editor}{\bibinfo{person}{Yaser Al-Onaizan}, \bibinfo{person}{Mohit Bansal}, {and} \bibinfo{person}{Yun-Nung Chen}} (Eds.). \bibinfo{publisher}{Association for Computational Linguistics}, \bibinfo{address}{Miami, Florida, USA}, \bibinfo{pages}{11709--11724}.
\newblock
\href{https://doi.org/10.18653/v1/2024.findings-emnlp.685}{doi:\nolinkurl{10.18653/v1/2024.findings-emnlp.685}}


\bibitem[Shannon(1948)]%
        {Shannon1948}
\bibfield{author}{\bibinfo{person}{Claude~E Shannon}.} \bibinfo{year}{1948}\natexlab{}.
\newblock \showarticletitle{A mathematical theory of communication}.
\newblock \bibinfo{journal}{\emph{The Bell system technical journal}} \bibinfo{volume}{27}, \bibinfo{number}{3} (\bibinfo{year}{1948}), \bibinfo{pages}{379--423}.
\newblock


\bibitem[Tang et~al\mbox{.}(2025)]%
        {survey_video_language_understanding_2025}
\bibfield{author}{\bibinfo{person}{Yunlong Tang}, \bibinfo{person}{Jing Bi}, \bibinfo{person}{Siting Xu}, {et~al\mbox{.}}} \bibinfo{year}{2025}\natexlab{}.
\newblock \showarticletitle{{Video Understanding with Large Language Models: A Survey}}.
\newblock \bibinfo{journal}{\emph{IEEE Transactions on Circuits and Systems for Video Technology}} (\bibinfo{date}{May} \bibinfo{year}{2025}), \bibinfo{pages}{1}.
\newblock
\href{https://doi.org/10.1109/TCSVT.2025.3566695}{doi:\nolinkurl{10.1109/TCSVT.2025.3566695}}


\bibitem[Wang and Ji(2024)]%
        {epistemic-uq2024}
\bibfield{author}{\bibinfo{person}{Hanjing Wang} {and} \bibinfo{person}{Qiang Ji}.} \bibinfo{year}{2024}\natexlab{}.
\newblock \showarticletitle{{Epistemic Uncertainty Quantification For Pre-trained Neural Network}}.
\newblock \bibinfo{journal}{\emph{ArXiv e-prints}} (\bibinfo{date}{April} \bibinfo{year}{2024}).
\newblock
\href{https://doi.org/10.48550/arXiv.2404.10124}{doi:\nolinkurl{10.48550/arXiv.2404.10124}}


\bibitem[Wang et~al\mbox{.}(2024)]%
        {Qwen2-VL}
\bibfield{author}{\bibinfo{person}{Peng Wang}, \bibinfo{person}{Shuai Bai}, \bibinfo{person}{Sinan Tan}, {et~al\mbox{.}}} \bibinfo{year}{2024}\natexlab{}.
\newblock \showarticletitle{{Qwen2-VL: Enhancing Vision-Language Model's Perception of the World at Any Resolution}}.
\newblock \bibinfo{journal}{\emph{ArXiv e-prints}} (\bibinfo{date}{Sept.} \bibinfo{year}{2024}).
\newblock
\showeprint{2409.12191}
\href{https://doi.org/10.48550/arXiv.2409.12191}{doi:\nolinkurl{10.48550/arXiv.2409.12191}}


\bibitem[Wang et~al\mbox{.}(2020)]%
        {wang2020minilm}
\bibfield{author}{\bibinfo{person}{Wenhui Wang}, \bibinfo{person}{Furu Wei}, \bibinfo{person}{Li Dong}, \bibinfo{person}{Hangbo Bao}, \bibinfo{person}{Nan Yang}, {and} \bibinfo{person}{Ming Zhou}.} \bibinfo{year}{2020}\natexlab{}.
\newblock \bibinfo{title}{MiniLM: Deep Self-Attention Distillation for Task-Agnostic Compression of Pre-Trained Transformers}.
\newblock
\showeprint[arxiv]{2002.10957}~[cs.CL]


\bibitem[Wienholt et~al\mbox{.}(2025)]%
        {DSE2025}
\bibfield{author}{\bibinfo{person}{Patrick Wienholt}, \bibinfo{person}{Sophie Caselitz}, \bibinfo{person}{Robert Siepmann}, {et~al\mbox{.}}} \bibinfo{year}{2025}\natexlab{}.
\newblock \showarticletitle{{Hallucination Filtering in Radiology Vision-Language Models Using Discrete Semantic Entropy}}.
\newblock \bibinfo{journal}{\emph{ArXiv e-prints}} (\bibinfo{date}{Oct.} \bibinfo{year}{2025}).
\newblock
\href{https://doi.org/10.48550/arXiv.2510.09256}{doi:\nolinkurl{10.48550/arXiv.2510.09256}}


\bibitem[Yang et~al\mbox{.}(2025)]%
        {Qwen3Yang2025May}
\bibfield{author}{\bibinfo{person}{An Yang}, \bibinfo{person}{Anfeng Li}, \bibinfo{person}{Baosong Yang}, {et~al\mbox{.}}} \bibinfo{year}{2025}\natexlab{}.
\newblock \showarticletitle{{Qwen3 Technical Report}}.
\newblock \bibinfo{journal}{\emph{ArXiv e-prints}} (\bibinfo{date}{May} \bibinfo{year}{2025}).
\newblock
\href{https://doi.org/10.48550/arXiv.2505.09388}{doi:\nolinkurl{10.48550/arXiv.2505.09388}}


\bibitem[Yin et~al\mbox{.}(2024)]%
        {survey_multimodal_llms_2025}
\bibfield{author}{\bibinfo{person}{Shukang Yin}, \bibinfo{person}{Chaoyou Fu}, \bibinfo{person}{Sirui Zhao}, {et~al\mbox{.}}} \bibinfo{year}{2024}\natexlab{}.
\newblock \showarticletitle{{A survey on multimodal large language models}}.
\newblock \bibinfo{journal}{\emph{National Science Review}} \bibinfo{volume}{11}, \bibinfo{number}{12} (\bibinfo{date}{Dec.} \bibinfo{year}{2024}), \bibinfo{pages}{nwae403}.
\newblock
\showISSN{2095-5138}
\href{https://doi.org/10.1093/nsr/nwae403}{doi:\nolinkurl{10.1093/nsr/nwae403}}


\bibitem[Zhang et~al\mbox{.}(2024)]%
        {RadFlag2024}
\bibfield{author}{\bibinfo{person}{Serena Zhang}, \bibinfo{person}{Sraavya Sambara}, \bibinfo{person}{Oishi Banerjee}, \bibinfo{person}{Julian Acosta}, \bibinfo{person}{L.~John Fahrner}, {et~al\mbox{.}}} \bibinfo{year}{2024}\natexlab{}.
\newblock \showarticletitle{{RadFlag: A Black-Box Hallucination Detection Method for Medical Vision Language Models}}.
\newblock \bibinfo{journal}{\emph{ArXiv e-prints}} (\bibinfo{date}{Nov.} \bibinfo{year}{2024}).
\newblock
\href{https://doi.org/10.48550/arXiv.2411.00299}{doi:\nolinkurl{10.48550/arXiv.2411.00299}}


\end{thebibliography}

\end{document}